\documentclass[10pt,twocolumn,letterpaper]{article}

\usepackage{cvpr}              
\usepackage[table,xcdraw]{xcolor}
\usepackage{amsmath}
\usepackage{float}
\usepackage{multirow}
\usepackage{changepage}
\usepackage[most]{tcolorbox}
\usepackage{colortbl}
\usepackage{mdframed}
\usepackage{pifont}
\usepackage{framed}
\usepackage{color}
\usepackage{marvosym}
\usepackage{balance}

\definecolor{cvprblue}{rgb}{0.21,0.49,0.74}
\definecolor{gray}{rgb}{0.92, 0.92, 0.92}
\definecolor{yellow}{rgb}{1, 1, 0.7}
\definecolor{orange}{rgb}{1, 0.85, 0.7}
\definecolor{red}{rgb}{1, 0.7, 0.7}

\definecolor{mygray}{gray}{.95}

\newenvironment{formal}{%
  \MakeFramed{\advance\hsize-\width\FrameRestore}%
  \noindent\hspace{-4.55pt}%
  \begin{adjustwidth}{}{}%
  \vspace{2pt}\vspace{-1pt}%
}
{%
  \vspace{-1pt}\end{adjustwidth}\endMakeFramed%
}

\definecolor{formalshade}{RGB}{213,230,242}
\definecolor{myblue}{RGB}{0, 77, 128}
\definecolor{bubbles}{rgb}{0.91, 1.0, 1.0}

\usepackage[pagebackref,breaklinks,colorlinks,citecolor=cvprblue]{hyperref}
\usepackage{cuted}
\def\paperID{404} 
\def\confName{3DV\xspace}
\def\confYear{2025\xspace}

\title{XLD: A Cross-Lane Dataset for Benchmarking Novel Driving View Synthesis}

\author{Hao Li$^{1,2,}$\thanks{Equal contribution.}
\quad
Chenming Wu$^{2,*}$
\quad
Ming Yuan$^{2}$
\quad
Yan Zhang$^{2}$
\quad
Chen Zhao$^{2}$
\quad
Chunyu Song$^{2}$
\quad\\
Haocheng Feng$^{2}$ 
\quad
Errui Ding$^{2}$
\quad
Dingwen Zhang$^{1,}$\textsuperscript{\Letter}
\quad
Jingdong Wang$^{2}$
\vspace{0.2cm}
\\
{\normalsize $^{1}$ BRAIN Lab, NWPU \quad $^{2}$ Baidu VIS}
}

\begin{document}
\maketitle
\begin{abstract}
Comprehensive testing of autonomous systems through simulation is essential to ensure the safety of autonomous driving vehicles. This requires the generation of safety-critical scenarios that extend beyond the limitations of real-world data collection, as many of these scenarios are rare or rarely encountered on public roads.
However, evaluating most existing novel view synthesis (NVS) methods relies on sporadic sampling of image frames from the training data, comparing the rendered images with ground-truth images. Unfortunately, this evaluation protocol falls short of meeting the actual requirements in closed-loop simulations. Specifically, the true application demands the capability to render novel views that extend beyond the original trajectory (such as cross-lane views), which are challenging to capture in the real world.
To address this, this paper presents a synthetic dataset for novel driving view synthesis evaluation, which is specifically designed for autonomous driving simulations. This unique dataset includes testing images captured by deviating from the training trajectory by $1-4$ meters. It comprises six sequences that cover various times and weather conditions. Each sequence contains $450$ training images, $120$ testing images, and their corresponding camera poses and intrinsic parameters. Leveraging this novel dataset, we establish the first realistic benchmark for evaluating existing NVS approaches under front-only and multicamera settings. The experimental findings underscore the significant gap in current approaches, revealing their inadequate ability to fulfill the demanding prerequisites of cross-lane or closed-loop simulation.
Our dataset and code are released publicly on the project page: \href{https://3d-aigc.github.io/XLD}{https://3d-aigc.github.io/XLD}.
\end{abstract}    
\section{Introduction}

Autonomous driving (AD) simulation, which bridges the gap between the real and virtual worlds, is essential for testing and developing autonomous driving software in vehicles~\cite{berger2014engineering}. Research indicates that employing effective simulation methods can significantly expedite the evaluation of safety tests for autonomous driving, achieving a speedup of approximately $10^3$ to $10^5$ times faster than real-world testing~\cite{feng2023dense}. This compelling evidence underscores the importance of leveraging simulation to enhance the efficiency and effectiveness of autonomous driving development. However, the self-driving industry primarily conducts system testing using two approaches: log replay, which involves testing on pre-recorded real-world sensor data, and real-world driving, where new miles are driven to gather additional data for testing purposes~\cite{yang2023unisim}.
In closed-loop simulation, the vehicle must be free to respond to control commands within the simulation environment rather than strictly following the original trajectory from logs.
To promote the rapid advancement of end-to-end autonomous driving systems~\cite{zheng2024genad,hu2023planning}, designing a neural simulator for AD simulation~\cite{yang2023unisim,wu2023mars,wu2023mapnerf}, which can render photo-realistic images on novel views for closed-loop simulation and algorithm training, is in high demands.
The main scientific problems boil down to the 3D reconstruction~\cite{liu2018object}, and novel view synthesis (NVS)~\cite{yan2024street,wu2023mars}, which are also long-standing problems in computer vision and computer graphics. Traditional methods such as~\cite{schoenberger2016sfm,schoenberger2016mvs} have dominated the major deployment of 3D scene reconstruction for a long time. However, those reconstructed scenes cannot be directly used to produce photo-realistic novel views, thus imposing large restrictions on sensor simulations. As a result, the industrial bridges the sim-to-real gap by parametric and procedural modeling technique~\cite{tsirikoglou2017procedural} or human-involved creations.
With the recent rapid development of 3D implicit fields, such as neural radiance field (NeRF~\cite{mildenhall2021nerf}) and explicit primitive representations, i.e., 3D Gaussian Splatting (3DGS~\cite{kerbl3Dgaussians}), reconstructing a scene from a collection of images serves as the foundation of end-to-end autonomous driving simulation~\cite{he2024neural}. 
These techniques enable the rendering of high-quality and photorealistic images on novel views.

Presently, most approaches evaluate the performance of NVS results by splitting the dataset into training and testing sets. However, this strategy of splitting and sampling leads to an interpolation benchmark, which we argue is insufficient for evaluating whether the trained models can effectively render simulation-ready (\textit{i.e.,} cross-lane) and high-fidelity data for closed-loop simulations.
On the contrary, our proposed XLD dataset is a brand-new benchmark that evaluates the synthesis quality on a cross-lane view with additionally captured GT images.
Our dataset and benchmark focus on \textbf{assessing the NVS capability specifically for cameras in cross-lane scenes}. The primary objective is to evaluate the performance of cameras in generating accurate and realistic novel views in scenarios involving multiple lanes. 
Specifically, we introduce the XLD dataset, which encompasses the generation of $150\times 3$ rendered images for each scene. Furthermore, we evaluate novel cross-lane view synthesis by rendering $30$ images with deviations of $0$m, $1$m, $2$m, and $4$m from the training trajectory. Using the XLD dataset, we conduct a benchmark of leading methods, which are based on either NeRF or 3DGS, using well-established NVS metrics. Our benchmark results demonstrate that the proposed dataset offers a comprehensive evaluation benchmark tailored specifically to the requirements of closed-loop simulation. Furthermore, our benchmarking results reveal intriguing findings, emphasizing the value of the proposed dataset.
\section{Related Work}

\subsection{Autonomous Driving Simulation}
In the past few years, there has been a surge in the use of autonomous driving simulations~\cite{li2024choose}. These simulators are instrumental in validating planning and control mechanisms, producing educational and evaluative datasets, and significantly cutting down the time needed to perform these functions. The current landscape is dominated by two predominant categories of simulation tools: model-based and data-driven.
Model-based simulation platforms, such as PyBullet~\cite{coumans2016pybullet} MuJoCo~\cite{todorov2012mujoco}, AirSim~\cite{shah2018airsim} and CARLA~\cite{dosovitskiy2017carla}, utilize advanced computer graphics to replicate vehicles and their surroundings. However, the manual effort required to construct these models and program the vehicles' dynamics can be quite demanding and lengthy. Moreover, the visual output may sometimes fall short of the necessary realism, which can adversely affect the efficacy of perception systems when they are put into operation.

Previously, NVS heavily relied on conventional image processing techniques. For instance, Chaurasia et al.\cite{chaurasia2013depth} propose using depth synthesis from over-segmented graph structures. At the same time, AADS\cite{li2019aads} employs filtered and completed dense depth maps for warping novel view images through image stitching. 
A data-driven simulation platform VISTA~\cite{amini2022vista, amini2020learning} leverages datasets from the real world to create comprehensively labeled and photorealistic simulations. Recently, a wave of innovations has employed the NeRF method to simulate driving perspectives superficially. These new approaches excel in creating photorealistic images and have been shown to surpass traditional view synthesis algorithms in the realm of autonomous driving simulation.
Recent advances in neural novel view synthesis significantly accelerate the rapid development of the next-gen driving simulation, which exhibits superior expressiveness and flexibility compared to traditional methods. Our dataset and benchmark are specifically designed for those methods.

\subsection{NeRF-based NVS for Driving Simulation}
The introduction of neural radiance field (NeRF) revolutionized NVS by incorporating coordinate-based representation within multilayer perceptron (MLP) architectures, leading to significant performance improvements. Building upon NeRF, numerous subsequent works have further adapted these algorithms to fulfill requirements such as efficient training, anti-aliasing rendering, large-scale reconstruction, \textit{etc.}. 
InstantNGP~\cite{muller2022instant} proposes using a multi-resolution hash grid with a shallow MLP network to eliminate large MLP networks. 
Mip-NeRF~\cite{barron2021mip, barron2022mip} uses anti-aliased conical frustums instead of rays to reduce objectionable aliasing artifacts, which enables NeRF to represent fine details.
Zip-NeRF~\cite{barron2023zip} borrows the ideas from rendering and signal processing that combine Mip-NeRF with InstantNGP.
Nerfacto~\cite{tancik2023nerfstudio} integrates many advantages of existing methods to provide an all-in-one solution for NeRF training.
Block-NeRF~\cite{tancik2022block} tackles the reconstruction of large-scale urban scenes by division.
To handle dynamics, NSG~\cite{ost2021neural} decomposes dynamic scenes into scene graphs and learns a structured representation.
SUDS~\cite{turki2023suds} factorizes a large scene into three hash table data structures, encoding static, dynamic, and far-field radiance fields.
MARS~\cite{wu2023mars} is an instance-aware and modular simulator based on NeRF, which models dynamic foreground instances and static background environments separately.
UniSim~\cite{yang2023unisim} transforms a recorded log into a realistic closed-loop multi-sensor simulation, which incorporates dynamic object priors and utilizes a convolutional network to handle and complete unseen regions.
EmerNeRF~\cite{yang2023emernerf} employs a self-bootstrapping approach to simultaneously capture scene geometry, appearance, motion, and semantics, which enables comprehensive and synchronized modeling of these elements by stratifying scenes into static and dynamic fields.
UC-NeRF~\cite{cheng2023uc} addresses the challenge of under-calibrated multi-view novel view synthesis through layer-based color correction and virtual warping techniques.

\begin{figure*}[t]
    \centering
    \includegraphics[width=0.9\linewidth]{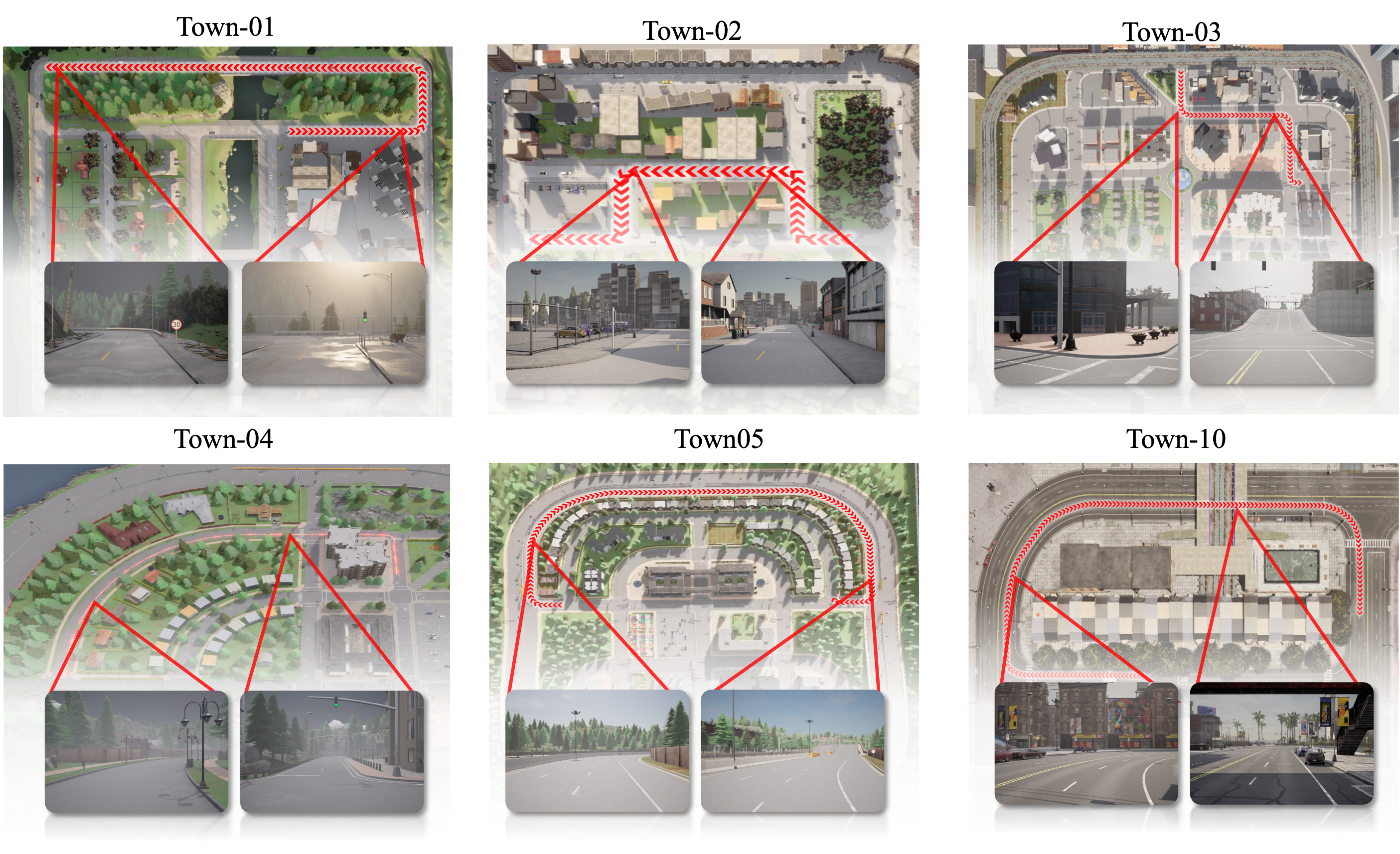}
    \caption{Our datasets encompass six distinct scenes, each involving the vehicle following an on-road trajectory. To generate training data for the cameras and LiDAR sensor, we sample $150$ waypoints along each trajectory. The trajectory is visually emphasized using the color red.}
    \label{fig:scenes}
\end{figure*}
\subsection{3DGS-based NVS for Driving Simulation}
Motivated by the NeRF-based methods and point-based differentiable rendering~\cite{yifan2019differentiable, aliev2020neural, ruckert2022adop, li2023dgnr}, 3D Gaussian Splatting (3DGS)~\cite{kerbl3Dgaussians} opens a new era with the leading advantages in explicit representation and real-time rendering capability. Within a concise timeframe, numerous methods~\cite{chen2023periodic, yan2024street} have emerged that focus on road scene reconstruction and NVS by leveraging the 3DGS representation. For instance, 
PVG~\cite{chen2023periodic, li2024VDG} introduces periodic vibration-based temporal dynamics to reconstruct dynamic urban scenes.
StreetGaussian~\cite{yan2024street} models the dynamic urban street environment as a collection of point clouds with semantic logits and 3D Gaussians, each associated with either a foreground vehicle or the background.
DrivingGaussian~\cite{zhou2023drivinggaussian} uses incremental static 3D Gaussians to represent the scene's static background. It also employs a composite dynamic Gaussian graph to handle multiple moving objects with LiDAR data.
HO-Gaussian~\cite{li2024ho} introduces a hybrid method that combines radiance fields with 3DGS representation, eliminating the requirement for point initialization in urban scene NVS.
HGS-Mapping~\cite{wu2024hgs} proposes a hybrid Gaussian representation specifically designed for performing online dense mapping in unbounded large-scale scenes.
GaussianPro~\cite{cheng2024gaussianpro} utilizes priors from reconstructed scene geometries and patch-matching techniques to generate precise Gaussians, leveraging the scene's existing structure.
DC-Gaussian~\cite{wang2024dc} introduces adaptive image decomposition for modeling reflections and occlusions. It incorporates illumination-aware obstruction modeling to handle reflections and occlusions under varying lighting conditions in urban scene novel view synthesis. Our dataset and benchmark specifically focus on evaluating the performance of neural-based driving simulation in NVS, particularly in cross-lane scenarios. A portion of the mentioned methods serve as our baselines, taking into account the code availability.
Moreover, generalizable 3D-GS methods such as PixelSplat~\cite{charatan2024pixelsplat}, Mvsplat~\cite{chen2024mvsplat}, and GGRt~\cite{li2024ggrt} also attempts to synthesize novel images within a well-trained generalizable feed-forward Gaussian networks.

\begin{figure*}[ht]
    \centering
    \includegraphics[width=0.9\linewidth]{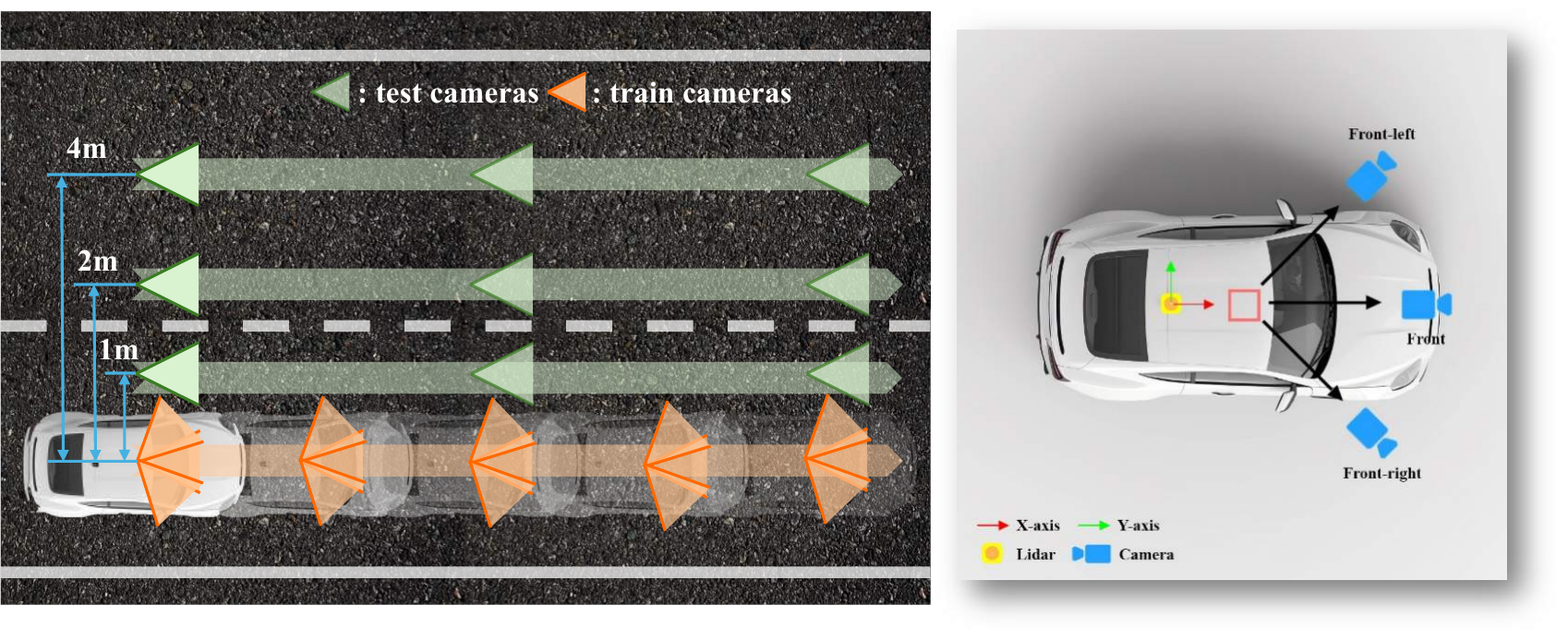}
    \caption{The composition of our training set and testing set. The training set consists of three RGB cameras ('front', 'left-front', and 'right-front') mounted on our vehicle lane. We sample image sequences along trajectories that run parallel to the vehicle's route for the test set. This encompasses four distinct test trajectories, each offset from the vehicle's trajectory by 0 meters, 1 meter, 2 meters, and 4 meters, respectively. The sampling interval for the test set is five times the sampling interval used for the training set. }
    \label{fig:cross_lane}
\end{figure*}

\subsection{Datasets in Autonomous Driving}

In autonomous driving training and benchmarking, there are many datasets available. For example, KITTI~\cite{geiger2012we}. KITTI-360~\cite{liao2022kitti}, vKITTI~\cite{cabon2020virtual}, CityScapes~\cite{cordts2016cityscapes}, Mapillary~\cite{neuhold2017mapillary}, ApolloScape~\cite{huang2018apolloscape}, Waymo Open Dataset~\cite{sun2020scalability}. nuScenes~\cite{caesar2020nuscenes}, Argoverse~\cite{chang2019argoverse} and Argoverse 2~\cite{wilson2023argoverse}, BDD100K~\cite{yu2020bdd100k}, OpenLane-V2~\cite{wang2024openlane}, LiDAR-CS~\cite{fang2024lidar}, \textit{etc.}. Those previous works have laid the groundwork for research and development in autonomous driving algorithms. A comprehensive survey of datasets related to autonomous driving refers to~\cite{liu2024survey}. More recently, a newly proposed Hierarchical 3D-GS~\cite{kerbl2024hierarchical} showcases the scalability of 3D-GS followed by a short scene with parallel trajectories to train and render. However, none of these existing researches have specifically addressed the evaluation of novel view synthesis techniques tailored for autonomous driving simulation, particularly in terms of their ability to meet the high demands of cross-lane capability.

\section{Dataset}

Distinguishing our dataset from previous datasets (both real-world and synthetic) that typically capture only a single road trajectory, ours additionally captures multiple parallel trajectories. 
To this end, we need to generate cross-lane data in the created worlds; here, we utilize Carla~\cite{dosovitskiy2017carla}, an autonomous driving simulator platform built on Unreal Engine. 

\subsection{Sensor Setup}
To meet the needs of most NeRF and 3D-GS algorithms, the sensors used to capture data include three color cameras (\textit{i.e.} `left-front', `front', `right-front') and one 3D laser scanner. The spatial relationships between sensors and the vehicle are fixed as shown in Fig.~\ref{fig:cross_lane}.
All three RGB cameras share identical intrinsic, lens parameters. Specifically, they have a resolution of $1920\times1280$, a field of view (FOV) of \(49.5^{\circ} \times 36.7^{\circ}\), a sensitivity of ISO 100, and a shutter speed of 5ms. The 3D laser scanner features 64 laser beams, a scanning range of 60 meters, and laser beam angles ranging from $-30^\circ$ to $30^\circ$ in the vertical direction. 

\subsection{Data Generation}
Our simulation environment consists of six scenes (``Town01," ``Town02," ``Town03," ``Town04," ``Town05," and ``Town10", all of them are under CC-BY License provided by~\cite{dosovitskiy2017carla}) with various weather conditions, such as sunshine and rain, closely resembling real-world settings. An overview of all the scenes and the trajectories is shown in Fig.~\ref{fig:scenes}.
For example, the training set samples 150 times in one scene to capture the images of three cameras, LiDAR points, and the vehicle's extrinsic information. The vehicle's forward distance between the two sample points is 2 meters.
For the evaluation process, in contrast to datasets captured in real-world scenarios, our cameras are configured with identical parameters, ensuring uniform image quality across all cameras.
Therefore, we only test the novel-view-synthesis in the view of \textit{front} camera, which includes four groups. These images are aligned parallel to the training set, with each group exhibiting a progressive deviation of 0m, 1m, 2m, and 4m along the y-axis in vehicle coordinate, as shown in Fig.~\ref{fig:cross_lane}. Additionally, a few novel-view-synthesis for AD methods like GaussianPro~\cite{cheng2024gaussianpro}, MARS~\cite{wu2023mars}, and UC-NeRF~\cite{cheng2023uc} need annotated sky masks to split the scene into foreground-sky and model the color compensation separately. We employ a pre-trained SegFormer~\cite{xie2021segformer} to effectively infer semantic segmentation masks and extract sky masks from them.

\section{Benchmark}

\begin{table*}[!htbp]
\centering
\footnotesize
\caption{Results on our proposed dataset with the different offsets using \emph{front-only} camera. $\uparrow$: higher is better, $\downarrow$: lower is better. The \colorbox{red}{red}, \colorbox{orange}{orange}, and \colorbox{yellow}{yellow} colors respectively denote the best, the second best, and the third best results.}
\label{tab:main_front}
\begin{tabular}{l|ccc|ccc|ccc|ccc}
\toprule 
Method &\multicolumn{3}{c|}{\textit{w/o} Offset} & \multicolumn{3}{c|}{Offset-1m} & \multicolumn{3}{c|}{Offset-2m} & \multicolumn{3}{c}{Offset-4m} \\
& $\uparrow$PSNR & $\uparrow$SSIM & $\downarrow$LPIPS & $\uparrow$PSNR & $\uparrow$SSIM & $\downarrow$LPIPS & $\uparrow$PSNR & $\uparrow$SSIM & $\downarrow$LPIPS & $\uparrow$PSNR & $\uparrow$SSIM & $\downarrow$LPIPS \\ \midrule

\multicolumn{4}{l}{\small\textit{- NeRF-based}}\\
\hspace{1em}Instant-NGP~\cite{muller2022instant}\xspace 
& 29.76 & 0.894 & 0.253
& 23.44 & 0.814 & 0.346
& 22.37 & 0.790 & 0.386 
& 20.95 & 0.768 & 0.443 \\
\hspace{1em}UC-NeRF~\cite{cheng2023uc}\xspace 
& \cellcolor{orange}35.95 & \cellcolor{orange}0.936 & 0.311
& \cellcolor{red}30.07 & \cellcolor{red}0.896 & 0.355
& \cellcolor{yellow}25.17 & \cellcolor{orange}0.863 & 0.367 
& 22.89 & 0.797 & 0.420 \\
\hspace{1em}MARS~\cite{wu2023mars}\xspace 
& 30.21 & 0.873 & \cellcolor{orange}0.146 
& \cellcolor{yellow}27.40 & 0.851 & \cellcolor{orange}0.169
& 24.95 & 0.847 & \cellcolor{orange}0.194
& \cellcolor{orange}23.29 & 0.818 & \cellcolor{orange}0.235 \\
\hspace{1em}NeRFacto~\cite{tancik2023nerfstudio}\xspace 
& 27.39 & 0.888 & 0.252 
& 23.49 & 0.824 & 0.314 
& 21.64 & 0.786 & 0.379 
& 20.82 & 0.769 & 0.412 \\
\hspace{1em}EmerNeRF~\cite{yang2023emernerf}\xspace 
& \cellcolor{yellow}31.76 & 0.907 & \cellcolor{red}0.126 
& \cellcolor{orange}28.66 & 0.878 & \cellcolor{red}0.150 
& \cellcolor{red}26.05 & 0.852 & \cellcolor{red}0.182 
& \cellcolor{red}24.80 & \cellcolor{yellow}0.837 & \cellcolor{red}0.203 \\
\midrule

\multicolumn{4}{l}{\small\textit{- Gaussian-based}} \\
\hspace{1em}3DGS~\cite{kerbl3Dgaussians}\xspace  
& 30.87 & 0.916 & 0.274 
& 23.26 & 0.873 & 0.334 
& 22.01 & 0.829 & 0.396
& 19.17 & 0.768 & 0.460\\
\hspace{1em}PVG~\cite{chen2023periodic}\xspace 
& \cellcolor{red}37.78 & \cellcolor{red}0.960 & \cellcolor{yellow}0.189
& 26.84 & \cellcolor{yellow}0.882 & \cellcolor{yellow}0.296
& 24.42 & \cellcolor{yellow}0.854 & \cellcolor{yellow}0.335
& \cellcolor{yellow}23.17 & \cellcolor{orange}0.841 & \cellcolor{yellow}0.353\\
\hspace{1em}GaussianPro~\cite{cheng2024gaussianpro}\xspace 
& 31.62 & \cellcolor{yellow}0.919 & 0.263
& 22.61 & 0.856 & 0.338 
& 21.26 & 0.819 & 0.383
& 18.75 & 0.772 & 0.445\\
\hspace{1em}DC-Gaussian~\cite{wang2024dc}\xspace
& 31.29 & 0.919 & 0.264 
& 26.82 & \cellcolor{orange}0.884 & 0.298
& \cellcolor{orange}25.24 & \cellcolor{red}0.871 & 0.319 
& 22.90 & \cellcolor{red}0.844 & 0.360 \\
\bottomrule
\end{tabular}
\end{table*}

\begin{table*}[t]
\footnotesize
\caption{Results on our proposed dataset with the different offsets using \emph{left-front, front, right-front} cameras. $\uparrow$: higher is better, $\downarrow$: lower is better. The \colorbox{red}{red}, \colorbox{orange}{orange}, and \colorbox{yellow}{yellow} colors respectively denote the best, the second best, and the third best results.}
\label{tab:main_3_cam}
\begin{tabular}{l|ccc|ccc|ccc|ccc}
\toprule 
Method &\multicolumn{3}{c|}{\textit{w/o} Offset} & \multicolumn{3}{c|}{Offset-1m} & \multicolumn{3}{c|}{Offset-2m} & \multicolumn{3}{c}{Offset-4m} \\
& $\uparrow$PSNR & $\uparrow$SSIM & $\downarrow$LPIPS & $\uparrow$PSNR & $\uparrow$SSIM & $\downarrow$LPIPS & $\uparrow$PSNR & $\uparrow$SSIM & $\downarrow$LPIPS & $\uparrow$PSNR & $\uparrow$SSIM & $\downarrow$LPIPS \\ \midrule

\multicolumn{4}{l}{\small\textit{- NeRF-based}}\\
\hspace{1em}Instant-NGP~\cite{muller2022instant}\xspace 
& 29.24 & 0.888 & 0.262
& 23.52 & 0.847 & 0.344
& 22.39 & 0.815 & 0.382
& 21.05 & 0.783 & 0.428 \\
\hspace{1em}UC-NeRF~\cite{cheng2023uc}\xspace 
& 33.12 & 0.912 & 0.360
& \cellcolor{red}30.07 & \cellcolor{red}0.896 & 0.355
& \cellcolor{red}28.81 & \cellcolor{red}0.881 & 0.373
& \cellcolor{red}26.87 & \cellcolor{red}0.870 & 0.421 \\
\hspace{1em}MARS~\cite{wu2023mars}\xspace 
& \cellcolor{yellow}31.37 & 0.887 & \cellcolor{orange}0.151
& \cellcolor{yellow}29.28 & 0.874 & \cellcolor{orange}0.157
& \cellcolor{yellow}28.54 & 0.869 & \cellcolor{orange}0.165
& \cellcolor{yellow}26.49 & 0.847 & \cellcolor{orange}0.193 \\
\hspace{1em}NeRFacto~\cite{tancik2023nerfstudio}\xspace 
& 29.39 & 0.890 & \cellcolor{yellow}0.246
& 22.91 & 0.850 & \cellcolor{yellow}0.307 
& 22.26 & 0.809 & 0.348
& 21.03 & 0.779 & 0.393 \\
\hspace{1em}EmerNeRF~\cite{yang2023emernerf}\xspace 
& \cellcolor{orange}31.51 & 0.894 & \cellcolor{red}0.146
& \cellcolor{orange}29.41 & 0.878 & \cellcolor{red}0.152
& \cellcolor{orange}28.66 & \cellcolor{orange}0.873 & \cellcolor{red}0.160
& \cellcolor{orange}26.62 & \cellcolor{yellow}0.851 & \cellcolor{red}0.188 \\
\midrule

\multicolumn{4}{l}{\small\textit{- Gaussian-based}} \\
\hspace{1em}3DGS~\cite{kerbl3Dgaussians}\xspace  
& 29.74 & \cellcolor{orange}0.914 & 0.312
& 22.08 & 0.842 & 0.359
& 21.34 & 0.824 & 0.402
& 19.47 & 0.796 & 0.441 \\
\hspace{1em}PVG~\cite{chen2023periodic}\xspace 
& \cellcolor{red}33.33 & \cellcolor{red}0.933 & 0.256
& 26.62 & \cellcolor{yellow}0.878 & 0.318
& 25.50 & 0.870 & 0.332
& 23.35 & 0.849 & 0.360 \\
\hspace{1em}GaussianPro~\cite{cheng2024gaussianpro}\xspace 
& 28.13 & 0.889 & 0.321
& 21.90 & 0.839 & 0.379
& 20.85 & 0.822 & 0.404
& 19.36 & 0.795 & 0.443 \\
\hspace{1em}DC-Gaussian~\cite{wang2024dc}\xspace
& 30.23 & \cellcolor{yellow}0.912 & 0.271
& 26.74 & \cellcolor{orange}0.883 & 0.315
& 25.53 & \cellcolor{yellow}0.872 & \cellcolor{yellow}0.329
& 23.53 & \cellcolor{orange}0.860 & \cellcolor{yellow}0.357 \\
\bottomrule
\end{tabular}
\end{table*}

\subsection{Benchmarking Environment}
To comprehensively evaluate the performance and computational efficiency of the assessment methods, we conducted a series of experiments using an NVIDIA Tesla V100 16GB GPU. We benchmarked the selected methods across five different cities that include various driving scenarios. Our findings are presented through both qualitative and quantitative analyses.

\begin{figure*}[t]
    \centering
    \includegraphics[width=1\linewidth]{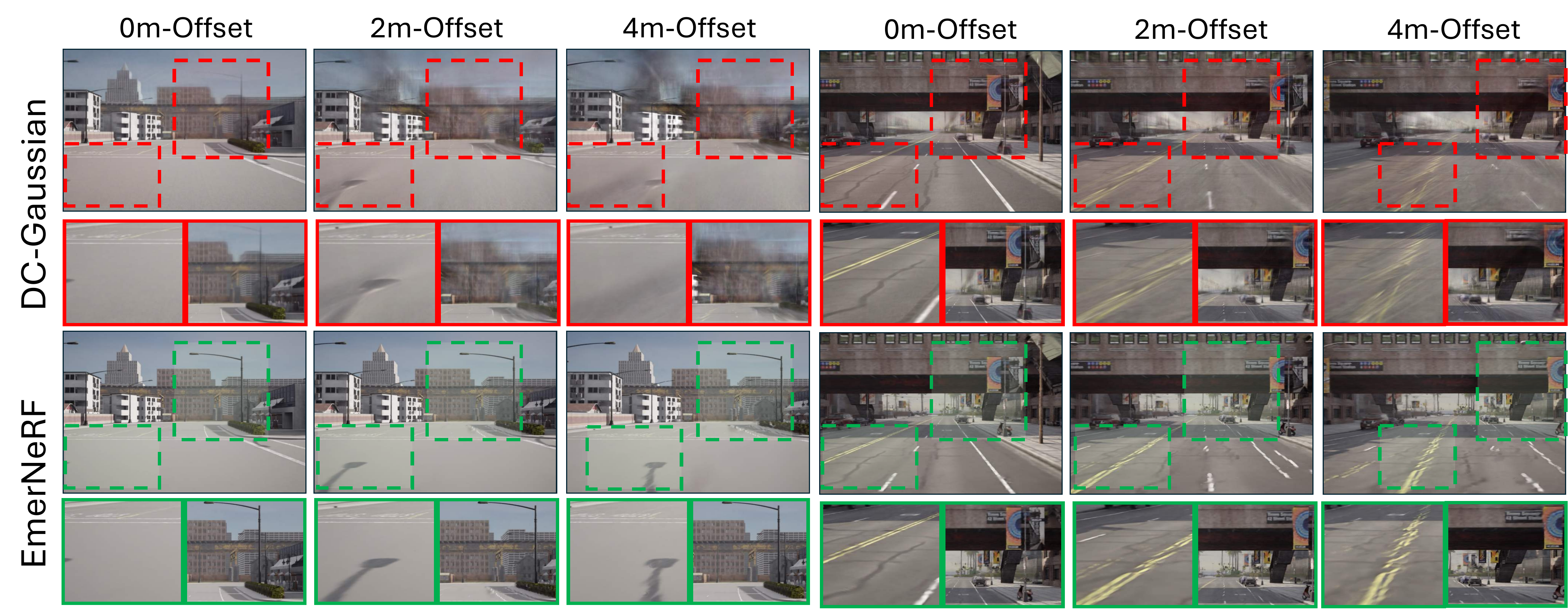}
    \caption{Visualization of the rendered images using DC-Gaussian and EmerNeRF with different offsets (\textit{i.e.} 0m, 2m, and 4m) in two scenes. The discriminated areas are highlighted, and the areas with better results are marked as \textcolor{green}{\(\square\)}, while worse results are marked as \textcolor{red}{\(\square\)}.
}
    \label{fig:finding_nerf_3dgs}
\end{figure*}

\subsection{Benchmarking Methods}

\noindent\textbf{InstantNGP~\cite{muller2022instant}:} We employ the Adam optimizer and maintain similar parameter settings as the original Instant-NGP implementation: the learning rate is $1\times 10^{-4}$, the number of feature dimensions per entry is $F=8$, the number of levels is $L=10$, and the hash tables is $2^4$. We train the model with 30,000 steps.

\noindent\textbf{Nerfacto~\cite{tancik2023nerfstudio}:} We use the implementation in~\cite{tancik2023nerfstudio} without pose refinement to test our benchmark. We employ the Adam optimizer with $1\times 10^{-3}$ learning rate. We train the model with 30,000 steps.

\noindent\textbf{MARS~\cite{wu2023mars}:} We inherit most of the parameter settings as the original MARS implementation. We employ the RAdam optimizer with $1\times 10^{-3}$ learning rate.  Since our scenes are static without moving objects, we disable \(\mathcal{L}_{sem}\), and the rest of the loss functions remain the same.
We train the model with 50,000 steps.

\noindent\textbf{UC-NeRF~\cite{cheng2023uc}:}  We inherit most parameter settings from the original UC-NeRF implementation. 
We employ the AdamW optimizer with $2.5\times 10^{-3}$ learning rate.  The weight of sky loss is set to \(2\times 10^{-3}\). We train the model with 40,000 steps.

\noindent\textbf{3DGS~\cite{kerbl3Dgaussians}:} We use the implementation of NerfStudio to evaluate our dataset. We employ the AdamW optimizer with $1\times 10^{-3}$ learning rate. For stability, we “warm up” the computation in lower resolution. Specifically, we start the optimization using 4 times smaller image resolution and then upsample twice after 500 and 1000 iterations. We train the model with 30,000 steps.

\noindent\textbf{PVG~\cite{chen2023periodic}:} We employ the Adam optimizer and maintain a similar learning rate for most parameters as the original PVG implementation. At the same time, we adjust the learning rate of the amplitude $\mathbf{A}$, opacity decaying $\mathbf{\beta}$ and opacity $\mathbf{O}$ to $3\times 10^{-5}$, $0.02$ and $0.005$ respectively. We train the model with 30,000 steps.

\noindent\textbf{GaussianPro~\cite{cheng2024gaussianpro}:} In alignment with the approach described in GaussianPro, our models are trained for 30,000 iterations across all scenes following GaussianPro’s training schedule and hyper-parameters. The interval step of the progressive propagation strategy is set to 50 where propagation is performed 3 times. The threshold \(\sigma\) of the absolute relative difference is set to 0.8. We set \(\beta=0.001\) and \(\gamma=0.001\) for the planar loss.

\noindent\textbf{DC-Gaussian~\cite{wang2024dc}:} To align the performance described in DC-Gaussian, we set 
loss coefficient 0.001 for both photometric and sky losses. Sky loss is the same as UC-NeRF. 

\subsection{Used Metrics}
We adopt the evaluation criteria employed by the methods mentioned above, which comprise Peak Signal-to-Noise Ratio (PSNR), Structural Similarity (SSIM), and Learned Perceptual Image Patch Similarity (LPIPS) as our evaluation metrics. Furthermore, we ensure transparency and clarity by describing the experimental framework employed to compare the methods.

\subsection{Experimental Results}
We evaluate the methods mentioned below on our dataset using novel-view-synthesis metrics. These methods are trained in two modes: (1) front-only mode (Tab. ~\ref{tab:main_front}) utilizes images captured solely by the front camera; (2) multi-camera mode (Tab. ~\ref{tab:main_3_cam}) incorporates images from all three cameras. We conduct separate evaluations on trajectories with 0m, 1m, 2m, and 4m offsets. Our evaluation encompasses qualitative and quantitative experiments, ensuring a comprehensive analysis of the methods' performance.
Furthermore, it is worth noting that EmerNeRF~\cite{yang2023emernerf} demonstrates the highest performance in the cross-lane dataset, with an average PSNR of 26.50 dB. We attribute this superior performance to its inherent self-supervised scene decomposition and positional embedding decomposition capability.
We place detailed experimental results in Tab.~\ref{tab:main_front} and Tab. \ref{tab:main_3_cam}. For more detailed results, please refer to the supplementary material and our webpage.

\begin{figure*}[t]
    \centering
    \includegraphics[width=1\linewidth]{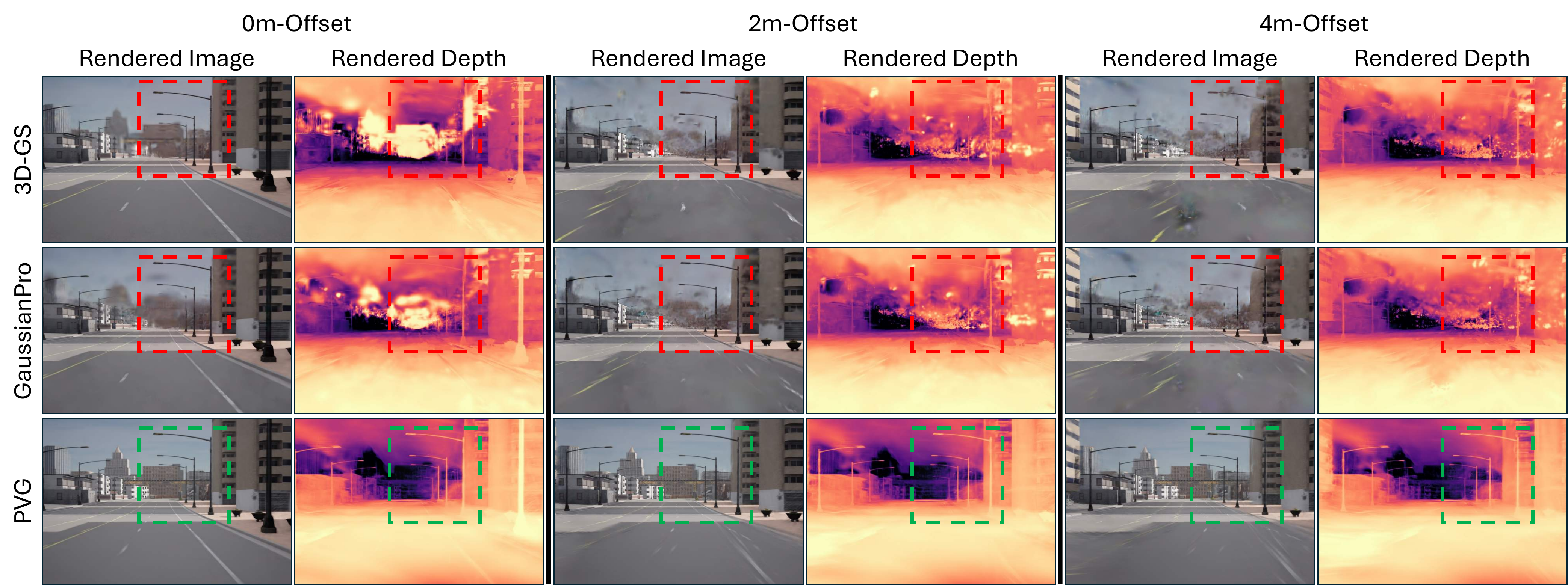}
    \caption{Visualization of the rendered images and depth maps using 3D-GS~\cite{kerbl3Dgaussians}, GaussianPro~\cite{cheng2024gaussianpro}, and PVG~\cite{chen2023periodic} with different offsets (\textit{i.e.} 0m, 2m, and 4m) in two scenes. The discriminated areas are highlighted, and the areas with better results are marked as '\textcolor{green}{\(\square\)}'.
}
    \label{fig:finding_3dgs_bg}
\end{figure*}

\section{Findings}
\begin{formal}
\textbf{NeRFs perform better than 3D-GS averagely}
\end{formal}

Lately, there has been a shift in research interest within the community, transitioning from NeRF towards 3D-GS. 3D-GS has achieved state-of-the-art results on datasets such as KITTI and Waymo. However, it is worth noting that these datasets primarily focus on evaluating interpolation images without considering significant offsets (\textit{i.e.} cross-lane NVS). Consequently, a crucial aspect of our dataset is to assess and compare the novel-view synthesis performance between NeRF and 3D-GS methods under challenging conditions involving substantial offsets.

As shown in Tab. \ref{tab:main_3_cam}, in the scenario with no offset, compared with EmerNeRF~\cite{yang2023emernerf}, PVG~\cite{chen2023periodic} achieves better performance by 0.2dB in PSNR metrics with 3-cameras setting and 1.83dB in PSNR metrics with 1-camera setting. In the scenario with 4m offset, EmerNeRF~\cite{yang2023emernerf} and UC-NeRF~\cite{cheng2023uc} show a much more powerful ability to synthesize images in cross-lane novel view. compared with the SOTA Gaussian method (\textit{i.e.} DC-Gaussian~\cite{wang2024dc}), they achieve leading performance by 1.90dB and 3.43dB in PSNR separately. The visualization results are shown in Fig. \ref{fig:finding_nerf_3dgs}.  

\begin{formal}
\textbf{Self-decomposition handles background better}
\end{formal}

In this comparison, we examine the self-decomposition method (PVG~\cite{chen2023periodic}) alongside traditional approaches (3D-GS~\cite{kerbl3Dgaussians} and GaussianPro~\cite{cheng2024gaussianpro}), the quantitative results are shown in Tab.~\ref{tab:finding_3dgs_bg}. The traditional method initializes 3D-GS primitives, denoted as \(\mathcal{G}=\left\{\left(\sigma_i, \boldsymbol{\mu}_i, \Sigma_i, c_i\right)\right\}_{i=1}^G\), using LiDAR point clouds and extends them to areas lacking geometric features through cloning and splitting operations. However, these techniques lack a geometry prior and heavily rely on photometric loss, resulting in an issue of over-fitting. This problem becomes evident in our benchmark, particularly in the context of cross-lane novel view synthesis, as illustrated in Fig.~\ref{fig:finding_3dgs_bg}.
\setlength{\tabcolsep}{2pt}
\begin{table}[b]
\centering
\scriptsize
\caption{Results on three 3D-GS methods with the different offsets with multi-cameras setting.}
\label{tab:finding_3dgs_bg}
\begin{tabular}{c|ccc|ccc|ccc}
\toprule
\multirow{2}{*}{Offset} & \multicolumn{3}{c|}{3D-GS} & \multicolumn{3}{c|}{GaussianPro} & \multicolumn{3}{c}{PVG} \\
                        & PSNR$\uparrow$    & SSIM$\uparrow$    & LPIPS$\downarrow$  & PSNR$\uparrow$      & SSIM$\uparrow$      & LPIPS$\downarrow$    & PSNR$\uparrow$   & SSIM$\uparrow$   & LPIPS$\downarrow$ \\ \midrule
0m                      & 27.23   & 0.892   & 0.278  & 26.97     & 0.890     & 0.283    & 34.08  & 0.948  & 0.219 \\
1m                      & 22.33   & 0.846   & 0.331  & 21.89     & 0.840     & 0.335    & 28.11  & 0.907  & 0.270 \\
2m                      & 21.15   & 0.825   & 0.355  & 20.76     & 0.819     & 0.362    & 26.63  & 0.894  & 0.289 \\
4m                      & 19.24   & 0.797   & 0.393  & 19.01     & 0.792     & 0.399    & 23.96  & 0.873  & 0.317 \\ \bottomrule
\end{tabular}
\end{table}

When not initialized by point clouds, the Gaussian points representing the background tend to overfit the training data to minimize rendering loss, often at the expense of positional accuracy. While they may appear satisfactory within the training trajectory, their performance degrades significantly when rendering images with offsets from the training trajectory. In Fig.~\ref{fig:finding_3dgs_bg}, it can be observed that the street lamps are accurately modeled in the rendered images without any offset. However, when the offset is increased to 2m and 4m, these lamps become obscured by the background Gaussian points, resulting in a substantial decline in performance. Specifically, there is a decrease in performance of -7.99dB and -7.69dB for 3D-GS~\cite{kerbl3Dgaussians} and GaussianPro~\cite{cheng2024gaussianpro}, respectively, as we move from 0 meters to 4 meters offset.

In contrast, the self-decomposition method, such as PVG~\cite{chen2023periodic}, incorporates frequency information into 3D-GS primitives to create a unified representation of dynamic objects and the background. These primitives are denoted as \(\mathcal{G}^k_t  =\{\tilde{\mu}^k_t, \Sigma^k, \widetilde{\alpha}^k_t, S^k\}_{k=1}^G\). Additionally, PVG employs a high-resolution environment cube map, denoted as $f_{s k y}(d)=c_{s k y}$, to effectively handle high-frequency details in the sky, where 
\(d\) represents the ray direction. These techniques enable the network to model the sky and avoid local minima accurately. As depicted in Fig.~\ref{fig:finding_3dgs_bg}, PVG produces depth maps that better represent the background, leading to more realistic images than other Gaussian methods. It achieves an impressive improvement of 4.72dB in PSNR compared to the baseline 3D-GS~\cite{kerbl3Dgaussians}, resulting in higher fidelity rendered images.

\begin{figure}[t]
    \centering
    \includegraphics[width=1\linewidth]{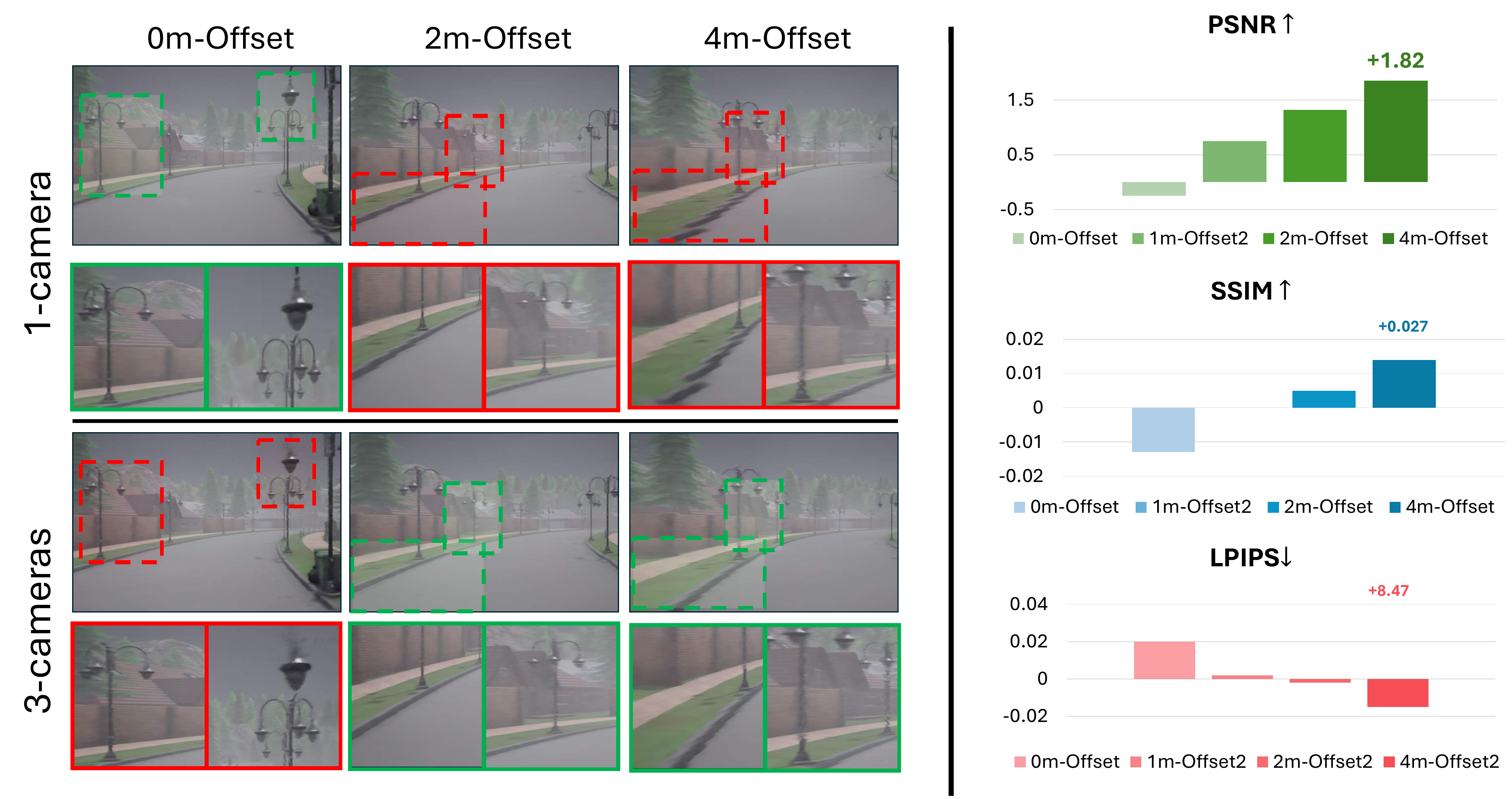}
    \caption{Use EmerNeRF~\cite{yang2023emernerf} as example. In the left column: we visualize the novel-view-synthesis results with different offsets and camera numbers, the discriminate areas are highlighted. In the right column, we show the performance improvement between 3-cameras and 1-camera settings using PSNR, SSIM, and LPIPS.
}
    \label{fig:finding_multi_cam}
\end{figure}

\begin{formal}
\textbf{Multi-camera benefits cross-lane NVS}
\end{formal}
Using EmerNeRF~\cite{yang2023emernerf} as an example, we compare the performance differences between multiple-cameras and front-camera settings shown in Fig. \ref{fig:finding_multi_cam}. 
NeRF with only front camera training performs better than the network using three cameras by 0.25db in PSNR metrics when rendering images without offsets. 
However, when it comes to an offset, we observe a trend that NeRF using multiple-camera training performs better than using only front-camera training, up to 1.82db PSNR improvement in the scenario of 4m offset.
These results indicate that using more cameras can significantly boost the performance of cross-lane novel view synthesis ability.

\begin{formal}
    \textbf{Geometric quality is key to cross-lane NVS}
\end{formal}
NeRF and 3D-GS methods have excellently performed on established autonomous driving datasets such as KITTI and Waymo. In these datasets, the primary function of the NVS methods is often to replay or interpolate existing views within the training trajectory rather than generating images in truly novel views.
However, our cross-lane novel view synthesis benchmark offers a distinct evaluation scenario that differs significantly from previous efforts validated on Waymo or KITTI datasets. This benchmark introduces novel challenges and evaluation criteria specifically designed to assess the performance of NVS methods when synthesizing images from viewpoints across different lanes. As a result, the evaluation of NeRF and 3D-GS methods in this benchmark provides unique insights beyond their performance on traditional datasets.

\begin{figure}[t]
    \centering
    \includegraphics[width=1\linewidth]{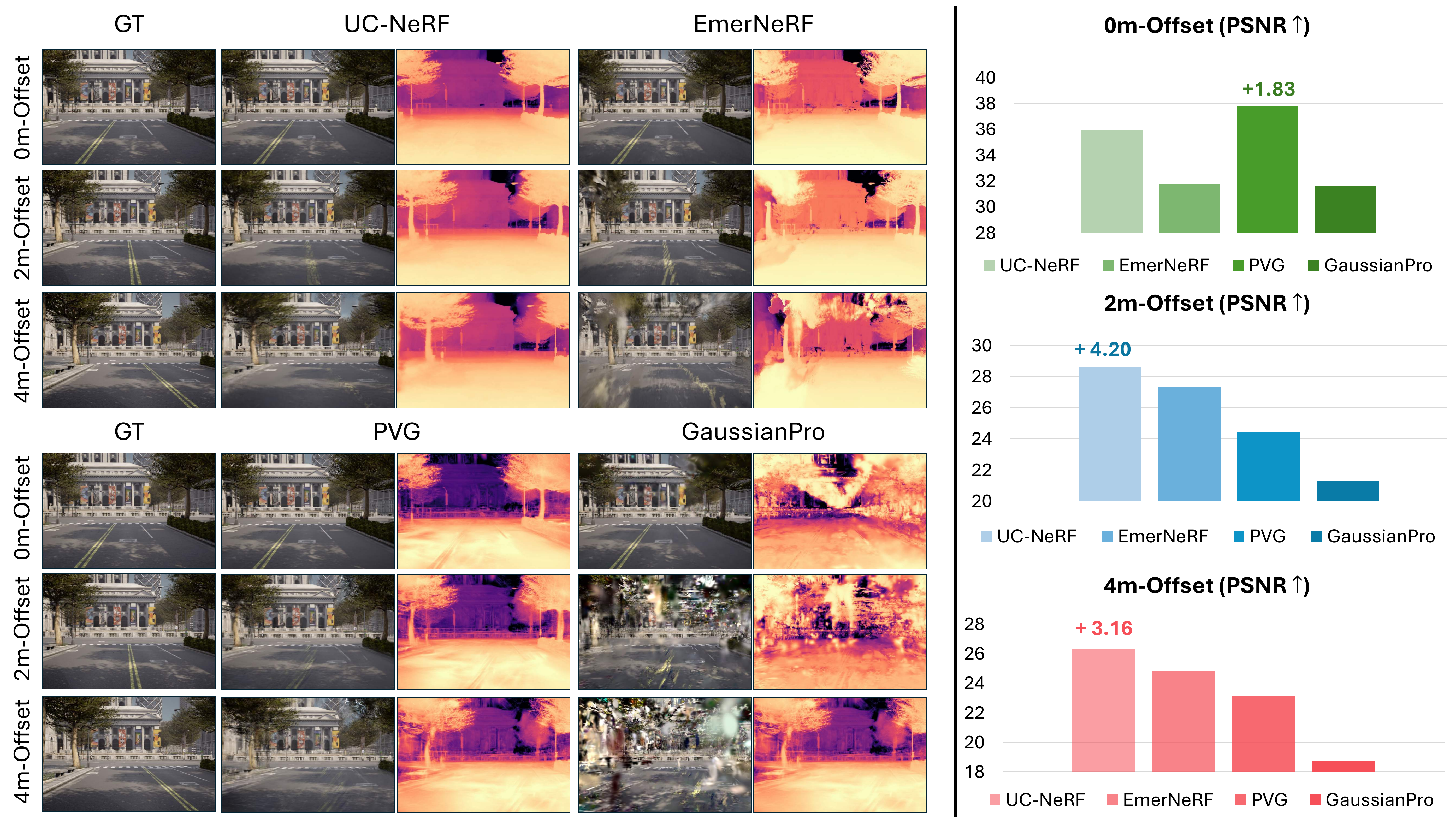}
    \caption{In the left column: we visualize the novel-view-synthesis results of four NeRF and Gaussian methods with different offsets. In the right column, we demonstrate the quantitative comparison between these four methods with different offsets using PSNR metrics.
}
    \label{fig:finding_geometry}
\end{figure}

As shown in the right column of Fig. \ref{fig:finding_geometry}, all the selected methods perform well in the scenarios without any offset, which is identical to the previous validation efforts, and PVG even achieves leading performance up to 37.78dB in PSNR. When increasing the offset to 4 meters, its performance drops significantly while UC-NeRF~\cite{cheng2023uc} and EmerNeRF~\cite{yang2023emernerf} surpass PVG by 3.16dB and 1.63dB in PSNR metric.

To this end, we visualize the rendered RGB images and depth maps with different methods, as shown in the left column of Fig. \ref{fig:finding_geometry}. 
UC-NeRF and EmerNeRF perform reasonable depth maps, properly distributing background buildings and sky. PVG successfully detaches the sky but performs poorly on the background buildings. GaussianPro overfits the training images and fails to estimate depth. These seriously affect the render performance in novel views with large offsets, where UC-NeRF, EmerNeRF, and PVG can render the rough structure of the scene in novel views, but GaussianPro fails.
These phenomena show that the NeRF / Gaussian methods need to achieve precise geometry reconstruction to obtain better performance in cross-lane NVS. 

\subsection{Discussion}

The results show that current approaches exhibit a substantial gap, indicating their limited capability to meet the rigorous requirements of cross-lane or closed-loop simulation. Future research endeavors can leverage our proposed dataset and benchmark to gauge how novel methods can advance toward achieving closed-loop simulation.

\noindent \textbf{Limitation.}~~Currently, our dataset has certain limitations as it was generated using the Unreal-based Carla simulator. We see our work represents an initial stride towards the accurate evaluation of novel driving view synthesis. As a future endeavor, we plan to curate a real-world dataset, similar to~\cite{ni2025paralane}, encompassing cross-lane ground truth data. This expansion will enhance the authenticity and applicability of our evaluation, advancing the field of novel driving view synthesis. 

\section{Conclusion}
In conclusion, this paper addresses the challenge that existing evaluation methods for NVS fall short of meeting the requirements of closed-loop simulations, which demand the capability to render views beyond the original trajectory. We introduce a unique driving view synthesis dataset and benchmark for autonomous driving simulations. This dataset includes testing images captured by deviating from the training trajectory, enabling the realistic evaluation of NVS approaches. Our dataset establishes a much-needed benchmark for advancing NVS techniques in closed-loop autonomous driving simulation.

\section*{Acknowledgements}
The authors would like to thank Yuanyuan Gao and Jingfeng Li for their assistance in preparing the experiments and the reviewers for their insightful feedback.

{
    \small
    \bibliographystyle{ieeenat_fullname}
    \bibliography{main}

\begin{thebibliography}{65}
\providecommand{\natexlab}[1]{#1}
\providecommand{\url}[1]{\texttt{#1}}
\expandafter\ifx\csname urlstyle\endcsname\relax
  \providecommand{\doi}[1]{doi: #1}\else
  \providecommand{\doi}{doi: \begingroup \urlstyle{rm}\Url}\fi

\bibitem[Aliev et~al.(2020)Aliev, Sevastopolsky, Kolos, Ulyanov, and Lempitsky]{aliev2020neural}
Kara-Ali Aliev, Artem Sevastopolsky, Maria Kolos, Dmitry Ulyanov, and Victor Lempitsky.
\newblock Neural point-based graphics.
\newblock In \emph{Proc. Eur. Conf. Comput. Vis. (ECCV)}, pages 696--712. Springer, 2020.

\bibitem[Amini et~al.(2020)Amini, Gilitschenski, Phillips, Moseyko, Banerjee, Karaman, and Rus]{amini2020learning}
Alexander Amini, Igor Gilitschenski, Jacob Phillips, Julia Moseyko, Rohan Banerjee, Sertac Karaman, and Daniela Rus.
\newblock Learning robust control policies for end-to-end autonomous driving from data-driven simulation.
\newblock \emph{IEEE Robot. Automat. Lett. (RA-L)}, 2020.

\bibitem[Amini et~al.(2022)Amini, Wang, Gilitschenski, Schwarting, Liu, Han, Karaman, and Rus]{amini2022vista}
Alexander Amini, Tsun-Hsuan Wang, Igor Gilitschenski, Wilko Schwarting, Zhijian Liu, Song Han, Sertac Karaman, and Daniela Rus.
\newblock Vista 2.0: An open, data-driven simulator for multimodal sensing and policy learning for autonomous vehicles.
\newblock In \emph{Proc. IEEE Int. Conf. Robot. Automat. (ICRA)}, 2022.

\bibitem[Barron et~al.(2021)Barron, Mildenhall, Tancik, Hedman, Martin-Brualla, and Srinivasan]{barron2021mip}
Jonathan~T Barron, Ben Mildenhall, Matthew Tancik, Peter Hedman, Ricardo Martin-Brualla, and Pratul~P Srinivasan.
\newblock Mip-nerf: A multiscale representation for anti-aliasing neural radiance fields.
\newblock In \emph{Proc.~of the IEEE/CVF Intl.~Conf.~on Computer Vision (ICCV)}, pages 5855--5864, 2021.

\bibitem[Barron et~al.(2022)Barron, Mildenhall, Verbin, Srinivasan, and Hedman]{barron2022mip}
Jonathan~T Barron, Ben Mildenhall, Dor Verbin, Pratul~P Srinivasan, and Peter Hedman.
\newblock Mip-nerf 360: Unbounded anti-aliased neural radiance fields.
\newblock In \emph{Proc. IEEE Conf. Comput. Vis. Pattern Recognit. (CVPR)}, pages 5470--5479, 2022.

\bibitem[Barron et~al.(2023)Barron, Mildenhall, Verbin, Srinivasan, and Hedman]{barron2023zip}
Jonathan~T Barron, Ben Mildenhall, Dor Verbin, Pratul~P Srinivasan, and Peter Hedman.
\newblock Zip-nerf: Anti-aliased grid-based neural radiance fields.
\newblock In \emph{Proc.~of the IEEE/CVF Intl.~Conf.~on Computer Vision (ICCV)}, pages 19697--19705, 2023.

\bibitem[Berger and Rumpe(2014)]{berger2014engineering}
Christian Berger and Bernhard Rumpe.
\newblock Engineering autonomous driving software.
\newblock \emph{arXiv preprint arXiv:1409.6579}, 2014.

\bibitem[Cabon et~al.(2020)Cabon, Murray, and Humenberger]{cabon2020virtual}
Yohann Cabon, Naila Murray, and Martin Humenberger.
\newblock Virtual kitti 2.
\newblock \emph{arXiv preprint arXiv:2001.10773}, 2020.

\bibitem[Caesar et~al.(2020)Caesar, Bankiti, Lang, Vora, Liong, Xu, Krishnan, Pan, Baldan, and Beijbom]{caesar2020nuscenes}
Holger Caesar, Varun Bankiti, Alex~H Lang, Sourabh Vora, Venice~Erin Liong, Qiang Xu, Anush Krishnan, Yu Pan, Giancarlo Baldan, and Oscar Beijbom.
\newblock nuscenes: A multimodal dataset for autonomous driving.
\newblock In \emph{Proc. IEEE Conf. Comput. Vis. Pattern Recognit. (CVPR)}, pages 11621--11631, 2020.

\bibitem[Chang et~al.(2019)Chang, Lambert, Sangkloy, Singh, Bak, Hartnett, Wang, Carr, Lucey, Ramanan, et~al.]{chang2019argoverse}
Ming-Fang Chang, John Lambert, Patsorn Sangkloy, Jagjeet Singh, Slawomir Bak, Andrew Hartnett, De Wang, Peter Carr, Simon Lucey, Deva Ramanan, et~al.
\newblock Argoverse: 3d tracking and forecasting with rich maps.
\newblock In \emph{Proc. IEEE Conf. Comput. Vis. Pattern Recognit. (CVPR)}, pages 8748--8757, 2019.

\bibitem[Charatan et~al.(2024)Charatan, Li, Tagliasacchi, and Sitzmann]{charatan2024pixelsplat}
David Charatan, Sizhe~Lester Li, Andrea Tagliasacchi, and Vincent Sitzmann.
\newblock pixelsplat: 3d gaussian splats from image pairs for scalable generalizable 3d reconstruction.
\newblock In \emph{Proc. IEEE Conf. Comput. Vis. Pattern Recognit. (CVPR)}, pages 19457--19467, 2024.

\bibitem[Chaurasia et~al.(2013)Chaurasia, Duchene, Sorkine-Hornung, and Drettakis]{chaurasia2013depth}
Gaurav Chaurasia, Sylvain Duchene, Olga Sorkine-Hornung, and George Drettakis.
\newblock Depth synthesis and local warps for plausible image-based navigation.
\newblock \emph{ACM Trans.~on Graphics (TOG)}, 32\penalty0 (3):\penalty0 1--12, 2013.

\bibitem[Chen et~al.(2023)Chen, Gu, Jiang, Zhu, and Zhang]{chen2023periodic}
Yurui Chen, Chun Gu, Junzhe Jiang, Xiatian Zhu, and Li Zhang.
\newblock Periodic vibration gaussian: Dynamic urban scene reconstruction and real-time rendering.
\newblock \emph{arXiv preprint arXiv:2311.18561}, 2023.

\bibitem[Chen et~al.(2024)Chen, Xu, Zheng, Zhuang, Pollefeys, Geiger, Cham, and Cai]{chen2024mvsplat}
Yuedong Chen, Haofei Xu, Chuanxia Zheng, Bohan Zhuang, Marc Pollefeys, Andreas Geiger, Tat-Jen Cham, and Jianfei Cai.
\newblock Mvsplat: Efficient 3d gaussian splatting from sparse multi-view images.
\newblock In \emph{Proc. Eur. Conf. Comput. Vis. (ECCV)}, 2024.

\bibitem[Cheng et~al.(2023)Cheng, Long, Yin, Wang, Wu, Ma, Wang, Chen, and Chen]{cheng2023uc}
Kai Cheng, Xiaoxiao Long, Wei Yin, Jin Wang, Zhiqiang Wu, Yuexin Ma, Kaixuan Wang, Xiaozhi Chen, and Xuejin Chen.
\newblock Uc-nerf: Neural radiance field for under-calibrated multi-view cameras.
\newblock In \emph{Proc.~of the Int.~Conf.~on Learning Representations (ICLR)}, 2023.

\bibitem[Cheng et~al.(2024)Cheng, Long, Yang, Yao, Yin, Ma, Wang, and Chen]{cheng2024gaussianpro}
Kai Cheng, Xiaoxiao Long, Kaizhi Yang, Yao Yao, Wei Yin, Yuexin Ma, Wenping Wang, and Xuejin Chen.
\newblock Gaussianpro: 3d gaussian splatting with progressive propagation.
\newblock In \emph{Proc.~of the Int.~Conf.~on Machine Learning (ICML)}, 2024.

\bibitem[Cordts et~al.(2016)Cordts, Omran, Ramos, Rehfeld, Enzweiler, Benenson, Franke, Roth, and Schiele]{cordts2016cityscapes}
Marius Cordts, Mohamed Omran, Sebastian Ramos, Timo Rehfeld, Markus Enzweiler, Rodrigo Benenson, Uwe Franke, Stefan Roth, and Bernt Schiele.
\newblock The cityscapes dataset for semantic urban scene understanding.
\newblock In \emph{Proc. IEEE Conf. Comput. Vis. Pattern Recognit. (CVPR)}, pages 3213--3223, 2016.

\bibitem[Coumans and Bai(2016)]{coumans2016pybullet}
Erwin Coumans and Yunfei Bai.
\newblock Pybullet, a python module for physics simulation for games, robotics and machine learning.
\newblock 2016.

\bibitem[Dosovitskiy et~al.(2017)Dosovitskiy, Ros, Codevilla, Lopez, and Koltun]{dosovitskiy2017carla}
Alexey Dosovitskiy, German Ros, Felipe Codevilla, Antonio Lopez, and Vladlen Koltun.
\newblock Carla: An open urban driving simulator.
\newblock In \emph{Proc.~of the Conf.~on Robot Learning (CoRL)}, 2017.

\bibitem[Fang et~al.(2024)Fang, Zhou, Zhao, Wu, Tang, Xu, and Zhang]{fang2024lidar}
Jin Fang, Dingfu Zhou, Jingjing Zhao, Chenming Wu, Chulin Tang, Cheng-Zhong Xu, and Liangjun Zhang.
\newblock Lidar-cs dataset: Lidar point cloud dataset with cross-sensors for 3d object detection.
\newblock In \emph{Proc. IEEE Int. Conf. Robot. Automat. (ICRA)}, pages 14822--14829, 2024.

\bibitem[Feng et~al.(2023)Feng, Sun, Yan, Zhu, Zou, Shen, and Liu]{feng2023dense}
Shuo Feng, Haowei Sun, Xintao Yan, Haojie Zhu, Zhengxia Zou, Shengyin Shen, and Henry~X Liu.
\newblock Dense reinforcement learning for safety validation of autonomous vehicles.
\newblock \emph{Nature}, 615\penalty0 (7953):\penalty0 620--627, 2023.

\bibitem[Geiger et~al.(2012)Geiger, Lenz, and Urtasun]{geiger2012we}
Andreas Geiger, Philip Lenz, and Raquel Urtasun.
\newblock Are we ready for autonomous driving? the kitti vision benchmark suite.
\newblock In \emph{Proc. IEEE Conf. Comput. Vis. Pattern Recognit. (CVPR)}, pages 3354--3361, 2012.

\bibitem[He et~al.(2024)He, Li, Sun, Han, Liu, Zheng, Wang, and Li]{he2024neural}
Lei He, Leheng Li, Wenchao Sun, Zeyu Han, Yichen Liu, Sifa Zheng, Jianqiang Wang, and Keqiang Li.
\newblock Neural radiance field in autonomous driving: A survey.
\newblock \emph{arXiv preprint arXiv:2404.13816}, 2024.

\bibitem[Hu et~al.(2023)Hu, Yang, Chen, Li, Sima, Zhu, Chai, Du, Lin, Wang, et~al.]{hu2023planning}
Yihan Hu, Jiazhi Yang, Li Chen, Keyu Li, Chonghao Sima, Xizhou Zhu, Siqi Chai, Senyao Du, Tianwei Lin, Wenhai Wang, et~al.
\newblock Planning-oriented autonomous driving.
\newblock In \emph{Proc. IEEE Conf. Comput. Vis. Pattern Recognit. (CVPR)}, pages 17853--17862, 2023.

\bibitem[Huang et~al.(2018)Huang, Cheng, Geng, Cao, Zhou, Wang, Lin, and Yang]{huang2018apolloscape}
Xinyu Huang, Xinjing Cheng, Qichuan Geng, Binbin Cao, Dingfu Zhou, Peng Wang, Yuanqing Lin, and Ruigang Yang.
\newblock The apolloscape dataset for autonomous driving.
\newblock In \emph{Proc.~of the IEEE/CVF Conf. on Computer Vision and Pattern Recognition Workshops}, pages 954--960, 2018.

\bibitem[Kerbl et~al.(2023)Kerbl, Kopanas, Leimk{\"u}hler, and Drettakis]{kerbl3Dgaussians}
Bernhard Kerbl, Georgios Kopanas, Thomas Leimk{\"u}hler, and George Drettakis.
\newblock 3d gaussian splatting for real-time radiance field rendering.
\newblock \emph{ACM Trans.~on Graphics (TOG)}, 2023.

\bibitem[Kerbl et~al.(2024)Kerbl, Meuleman, Kopanas, Wimmer, Lanvin, and Drettakis]{kerbl2024hierarchical}
Bernhard Kerbl, Andreas Meuleman, Georgios Kopanas, Michael Wimmer, Alexandre Lanvin, and George Drettakis.
\newblock A hierarchical 3d gaussian representation for real-time rendering of very large datasets.
\newblock \emph{ACM Trans.~on Graphics (TOG)}, 43\penalty0 (4):\penalty0 1--15, 2024.

\bibitem[Li et~al.(2024{\natexlab{a}})Li, Gao, Wu, Zhang, Dai, Zhao, Feng, Ding, Wang, and Han]{li2024ggrt}
Hao Li, Yuanyuan Gao, Chenming Wu, Dingwen Zhang, Yalun Dai, Chen Zhao, Haocheng Feng, Errui Ding, Jingdong Wang, and Junwei Han.
\newblock Ggrt: Towards generalizable 3d gaussians without pose priors in real-time.
\newblock In \emph{Proc. Eur. Conf. Comput. Vis. (ECCV)}, 2024{\natexlab{a}}.

\bibitem[Li et~al.(2024{\natexlab{b}})Li, Li, Zhang, Wu, Shi, Zhao, Feng, Ding, Wang, and Han]{li2024VDG}
Hao Li, Jingfeng Li, Dingwen Zhang, Chenming Wu, Jieqi Shi, Chen Zhao, Haocheng Feng, Errui Ding, Jingdong Wang, and Junwei Han.
\newblock Vdg: Vision-only dynamic gaussian for driving simulation.
\newblock \emph{arXiv preprint arXiv}, 2024{\natexlab{b}}.

\bibitem[Li et~al.(2019)Li, Pan, Zhang, Ren, Ma, Fang, Yan, Geng, Huang, Gong, et~al.]{li2019aads}
Wei Li, CW Pan, Rong Zhang, JP Ren, YX Ma, Jin Fang, FL Yan, QC Geng, XY Huang, HJ Gong, et~al.
\newblock Aads: Augmented autonomous driving simulation using data-driven algorithms.
\newblock \emph{Science robotics}, 2019.

\bibitem[Li et~al.(2024{\natexlab{c}})Li, Yuan, Zhang, Yan, Shen, Wang, and Yang]{li2024choose}
Yueyuan Li, Wei Yuan, Songan Zhang, Weihao Yan, Qiyuan Shen, Chunxiang Wang, and Ming Yang.
\newblock Choose your simulator wisely: A review on open-source simulators for autonomous driving.
\newblock \emph{IEEE Trans.~on Intelligent Vehicles}, 2024{\natexlab{c}}.

\bibitem[Li et~al.(2024{\natexlab{d}})Li, Wu, Zhang, and Zhu]{li2023dgnr}
Zhuopeng Li, Chenming Wu, Liangjun Zhang, and Jianke Zhu.
\newblock Dgnr: Density-guided neural point rendering of large driving scenes.
\newblock \emph{IEEE Transactions on Automation Science and Engineering}, 2024{\natexlab{d}}.

\bibitem[Li et~al.(2024{\natexlab{e}})Li, Zhang, Wu, Zhu, and Zhang]{li2024ho}
Zhuopeng Li, Yilin Zhang, Chenming Wu, Jianke Zhu, and Liangjun Zhang.
\newblock Ho-gaussian: Hybrid optimization of 3d gaussian splatting for urban scenes.
\newblock In \emph{Proc. Eur. Conf. Comput. Vis. (ECCV)}, 2024{\natexlab{e}}.

\bibitem[Liao et~al.(2022)Liao, Xie, and Geiger]{liao2022kitti}
Yiyi Liao, Jun Xie, and Andreas Geiger.
\newblock Kitti-360: A novel dataset and benchmarks for urban scene understanding in 2d and 3d.
\newblock \emph{IEEE Trans.~on Pattern Analalysis and Machine Intelligence (TPAMI)}, 45\penalty0 (3):\penalty0 3292--3310, 2022.

\bibitem[Liu et~al.(2018)Liu, Xia, Sun, Shen, Xu, Chen, Huang, and Xu]{liu2018object}
Ligang Liu, Xi Xia, Han Sun, Qi Shen, Juzhan Xu, Bin Chen, Hui Huang, and Kai Xu.
\newblock Object-aware guidance for autonomous scene reconstruction.
\newblock \emph{ACM Transactions on Graphics (TOG)}, 37\penalty0 (4):\penalty0 1--12, 2018.

\bibitem[Liu et~al.(2024)Liu, Yurtsever, Fossaert, Zhou, Zimmer, Cui, Zagar, and Knoll]{liu2024survey}
Mingyu Liu, Ekim Yurtsever, Jonathan Fossaert, Xingcheng Zhou, Walter Zimmer, Yuning Cui, Bare~Luka Zagar, and Alois~C Knoll.
\newblock A survey on autonomous driving datasets: Statistics, annotation quality, and a future outlook.
\newblock \emph{IEEE Trans.~on Intelligent Vehicles}, 2024.

\bibitem[Mildenhall et~al.(2021)Mildenhall, Srinivasan, Tancik, Barron, Ramamoorthi, and Ng]{mildenhall2021nerf}
Ben Mildenhall, Pratul~P Srinivasan, Matthew Tancik, Jonathan~T Barron, Ravi Ramamoorthi, and Ren Ng.
\newblock Nerf: Representing scenes as neural radiance fields for view synthesis.
\newblock \emph{Communications of the ACM}, 65\penalty0 (1):\penalty0 99--106, 2021.

\bibitem[M{\"u}ller et~al.(2022)M{\"u}ller, Evans, Schied, and Keller]{muller2022instant}
Thomas M{\"u}ller, Alex Evans, Christoph Schied, and Alexander Keller.
\newblock Instant neural graphics primitives with a multiresolution hash encoding.
\newblock \emph{ACM Trans.~on Graphics (TOG)}, 41\penalty0 (4):\penalty0 1--15, 2022.

\bibitem[Neuhold et~al.(2017)Neuhold, Ollmann, Rota~Bulo, and Kontschieder]{neuhold2017mapillary}
Gerhard Neuhold, Tobias Ollmann, Samuel Rota~Bulo, and Peter Kontschieder.
\newblock The mapillary vistas dataset for semantic understanding of street scenes.
\newblock In \emph{Proc.~of the IEEE/CVF Intl.~Conf.~on Computer Vision (ICCV)}, pages 4990--4999, 2017.

\bibitem[Ni et~al.(2025)Ni, Du, Hou, Wu, and Yang]{ni2025paralane}
Ziqian~Ni Ni, Sicong Du, Zhenghua Hou, Chenming Wu, and Sheng Yang.
\newblock Para-lane: Multi-lane dataset registering parallel scans for benchmarking novel view synthesis.
\newblock In \emph{Proc.~of the Intl.~Conf.~on {3D} Vision ({3D}V)}, 2025.

\bibitem[Ost et~al.(2021)Ost, Mannan, Thuerey, Knodt, and Heide]{ost2021neural}
Julian Ost, Fahim Mannan, Nils Thuerey, Julian Knodt, and Felix Heide.
\newblock Neural scene graphs for dynamic scenes.
\newblock In \emph{Proc. IEEE Conf. Comput. Vis. Pattern Recognit. (CVPR)}, pages 2856--2865, 2021.

\bibitem[R{\"u}ckert et~al.(2022)R{\"u}ckert, Franke, and Stamminger]{ruckert2022adop}
Darius R{\"u}ckert, Linus Franke, and Marc Stamminger.
\newblock Adop: Approximate differentiable one-pixel point rendering.
\newblock \emph{ACM Trans.~on Graphics (TOG)}, 41\penalty0 (4):\penalty0 1--14, 2022.

\bibitem[Sch\"{o}nberger and Frahm(2016)]{schoenberger2016sfm}
Johannes~Lutz Sch\"{o}nberger and Jan-Michael Frahm.
\newblock Structure-from-motion revisited.
\newblock In \emph{Conference on Computer Vision and Pattern Recognition (CVPR)}, 2016.

\bibitem[Sch\"{o}nberger et~al.(2016)Sch\"{o}nberger, Zheng, Pollefeys, and Frahm]{schoenberger2016mvs}
Johannes~Lutz Sch\"{o}nberger, Enliang Zheng, Marc Pollefeys, and Jan-Michael Frahm.
\newblock Pixelwise view selection for unstructured multi-view stereo.
\newblock In \emph{Proc. Eur. Conf. Comput. Vis. (ECCV)}, 2016.

\bibitem[Shah et~al.(2018)Shah, Dey, Lovett, and Kapoor]{shah2018airsim}
Shital Shah, Debadeepta Dey, Chris Lovett, and Ashish Kapoor.
\newblock Airsim: High-fidelity visual and physical simulation for autonomous vehicles.
\newblock In \emph{Field and Service Robotics: Results of the 11th International Conference}, pages 621--635. Springer, 2018.

\bibitem[Sun et~al.(2020)Sun, Kretzschmar, Dotiwalla, Chouard, Patnaik, Tsui, Guo, Zhou, Chai, Caine, et~al.]{sun2020scalability}
Pei Sun, Henrik Kretzschmar, Xerxes Dotiwalla, Aurelien Chouard, Vijaysai Patnaik, Paul Tsui, James Guo, Yin Zhou, Yuning Chai, Benjamin Caine, et~al.
\newblock Scalability in perception for autonomous driving: Waymo open dataset.
\newblock In \emph{Proc. IEEE Conf. Comput. Vis. Pattern Recognit. (CVPR)}, pages 2446--2454, 2020.

\bibitem[Tancik et~al.(2022)Tancik, Casser, Yan, Pradhan, Mildenhall, Srinivasan, Barron, and Kretzschmar]{tancik2022block}
Matthew Tancik, Vincent Casser, Xinchen Yan, Sabeek Pradhan, Ben Mildenhall, Pratul~P Srinivasan, Jonathan~T Barron, and Henrik Kretzschmar.
\newblock Block-nerf: Scalable large scene neural view synthesis.
\newblock In \emph{Proc. IEEE Conf. Comput. Vis. Pattern Recognit. (CVPR)}, pages 8248--8258, 2022.

\bibitem[Tancik et~al.(2023)Tancik, Weber, Ng, Li, Yi, Wang, Kristoffersen, Austin, Salahi, Ahuja, et~al.]{tancik2023nerfstudio}
Matthew Tancik, Ethan Weber, Evonne Ng, Ruilong Li, Brent Yi, Terrance Wang, Alexander Kristoffersen, Jake Austin, Kamyar Salahi, Abhik Ahuja, et~al.
\newblock Nerfstudio: A modular framework for neural radiance field development.
\newblock In \emph{ACM SIGGRAPH 2023 Conference Proceedings}, pages 1--12, 2023.

\bibitem[Todorov et~al.(2012)Todorov, Erez, and Tassa]{todorov2012mujoco}
Emanuel Todorov, Tom Erez, and Yuval Tassa.
\newblock Mujoco: A physics engine for model-based control.
\newblock In \emph{Proc.~of the IEEE/RSJ Intl.~Conf.~on Intelligent Robots and Systems (IROS)}, 2012.

\bibitem[Tsirikoglou et~al.(2017)Tsirikoglou, Kronander, Wrenninge, and Unger]{tsirikoglou2017procedural}
Apostolia Tsirikoglou, Joel Kronander, Magnus Wrenninge, and Jonas Unger.
\newblock Procedural modeling and physically based rendering for synthetic data generation in automotive applications.
\newblock \emph{arXiv preprint arXiv:1710.06270}, 2017.

\bibitem[Turki et~al.(2023)Turki, Zhang, Ferroni, and Ramanan]{turki2023suds}
Haithem Turki, Jason~Y Zhang, Francesco Ferroni, and Deva Ramanan.
\newblock Suds: Scalable urban dynamic scenes.
\newblock In \emph{Proc. IEEE Conf. Comput. Vis. Pattern Recognit. (CVPR)}, pages 12375--12385, 2023.

\bibitem[Wang et~al.(2024{\natexlab{a}})Wang, Li, Li, Chen, Sima, Liu, Wang, Jia, Wang, Jiang, et~al.]{wang2024openlane}
Huijie Wang, Tianyu Li, Yang Li, Li Chen, Chonghao Sima, Zhenbo Liu, Bangjun Wang, Peijin Jia, Yuting Wang, Shengyin Jiang, et~al.
\newblock Openlane-v2: A topology reasoning benchmark for unified 3d hd mapping.
\newblock In \emph{Proc.~of the Conference on Neural Information Processing Systems (NeurIPS)}, 2024{\natexlab{a}}.

\bibitem[Wang et~al.(2024{\natexlab{b}})Wang, Cheng, Lei, Wang, Yin, Lei, Long, and Lu]{wang2024dc}
Linhan Wang, Kai Cheng, Shuo Lei, Shengkun Wang, Wei Yin, Chenyang Lei, Xiaoxiao Long, and Chang-Tien Lu.
\newblock Dc-gaussian: Improving 3d gaussian splatting for reflective dash cam videos.
\newblock In \emph{Proc.~of the Conference on Neural Information Processing Systems (NeurIPS)}, 2024{\natexlab{b}}.

\bibitem[Wilson et~al.(2023)Wilson, Qi, Agarwal, Lambert, Singh, Khandelwal, Pan, Kumar, Hartnett, Pontes, et~al.]{wilson2023argoverse}
Benjamin Wilson, William Qi, Tanmay Agarwal, John Lambert, Jagjeet Singh, Siddhesh Khandelwal, Bowen Pan, Ratnesh Kumar, Andrew Hartnett, Jhony~Kaesemodel Pontes, et~al.
\newblock Argoverse 2: Next generation datasets for self-driving perception and forecasting.
\newblock \emph{arXiv preprint arXiv:2301.00493}, 2023.

\bibitem[Wu et~al.(2023{\natexlab{a}})Wu, Sun, Shen, and Zhang]{wu2023mapnerf}
Chenming Wu, Jiadai Sun, Zhelun Shen, and Liangjun Zhang.
\newblock Mapnerf: Incorporating map priors into neural radiance fields for driving view simulation.
\newblock In \emph{Proc.~of the IEEE/RSJ Intl.~Conf.~on Intelligent Robots and Systems (IROS)}, pages 7082--7088, 2023{\natexlab{a}}.

\bibitem[Wu et~al.(2024)Wu, Zhang, Zhang, Yuan, Tie, Wei, Xu, Zhao, Gan, and Ding]{wu2024hgs}
Ke Wu, Kaizhao Zhang, Zhiwei Zhang, Shanshuai Yuan, Muer Tie, Julong Wei, Zijun Xu, Jieru Zhao, Zhongxue Gan, and Wenchao Ding.
\newblock Hgs-mapping: Online dense mapping using hybrid gaussian representation in urban scenes.
\newblock \emph{arXiv preprint arXiv:2403.20159}, 2024.

\bibitem[Wu et~al.(2023{\natexlab{b}})Wu, Liu, Luo, Zhong, Chen, Xiao, Hou, Lou, Chen, Yang, et~al.]{wu2023mars}
Zirui Wu, Tianyu Liu, Liyi Luo, Zhide Zhong, Jianteng Chen, Hongmin Xiao, Chao Hou, Haozhe Lou, Yuantao Chen, Runyi Yang, et~al.
\newblock Mars: An instance-aware, modular and realistic simulator for autonomous driving.
\newblock In \emph{CAAI International Conference on Artificial Intelligence}, pages 3--15. Springer, 2023{\natexlab{b}}.

\bibitem[Xie et~al.(2021)Xie, Wang, Yu, Anandkumar, Alvarez, and Luo]{xie2021segformer}
Enze Xie, Wenhai Wang, Zhiding Yu, Anima Anandkumar, Jose~M Alvarez, and Ping Luo.
\newblock Segformer: Simple and efficient design for semantic segmentation with transformers.
\newblock In \emph{Proc.~of the Conference on Neural Information Processing Systems (NeurIPS)}, pages 12077--12090, 2021.

\bibitem[Yan et~al.(2024)Yan, Lin, Zhou, Wang, Sun, Zhan, Lang, Zhou, and Peng]{yan2024street}
Yunzhi Yan, Haotong Lin, Chenxu Zhou, Weijie Wang, Haiyang Sun, Kun Zhan, Xianpeng Lang, Xiaowei Zhou, and Sida Peng.
\newblock Street gaussians for modeling dynamic urban scenes.
\newblock In \emph{Proc. Eur. Conf. Comput. Vis. (ECCV)}, 2024.

\bibitem[Yang et~al.(2024)Yang, Ivanovic, Litany, Weng, Kim, Li, Che, Xu, Fidler, Pavone, et~al.]{yang2023emernerf}
Jiawei Yang, Boris Ivanovic, Or Litany, Xinshuo Weng, Seung~Wook Kim, Boyi Li, Tong Che, Danfei Xu, Sanja Fidler, Marco Pavone, et~al.
\newblock Emernerf: Emergent spatial-temporal scene decomposition via self-supervision.
\newblock In \emph{Proc.~of the Int.~Conf.~on Learning Representations (ICLR)}, 2024.

\bibitem[Yang et~al.(2023)Yang, Chen, Wang, Manivasagam, Ma, Yang, and Urtasun]{yang2023unisim}
Ze Yang, Yun Chen, Jingkang Wang, Sivabalan Manivasagam, Wei-Chiu Ma, Anqi~Joyce Yang, and Raquel Urtasun.
\newblock Unisim: A neural closed-loop sensor simulator.
\newblock In \emph{Proc. IEEE Conf. Comput. Vis. Pattern Recognit. (CVPR)}, pages 1389--1399, 2023.

\bibitem[Yifan et~al.(2019)Yifan, Serena, Wu, {\"O}ztireli, and Sorkine-Hornung]{yifan2019differentiable}
Wang Yifan, Felice Serena, Shihao Wu, Cengiz {\"O}ztireli, and Olga Sorkine-Hornung.
\newblock Differentiable surface splatting for point-based geometry processing.
\newblock \emph{ACM Trans.~on Graphics (TOG)}, 38\penalty0 (6):\penalty0 1--14, 2019.

\bibitem[Yu et~al.(2020)Yu, Chen, Wang, Xian, Chen, Liu, Madhavan, and Darrell]{yu2020bdd100k}
Fisher Yu, Haofeng Chen, Xin Wang, Wenqi Xian, Yingying Chen, Fangchen Liu, Vashisht Madhavan, and Trevor Darrell.
\newblock Bdd100k: A diverse driving dataset for heterogeneous multitask learning.
\newblock In \emph{Proc. IEEE Conf. Comput. Vis. Pattern Recognit. (CVPR)}, pages 2636--2645, 2020.

\bibitem[Zheng et~al.(2024)Zheng, Song, Guo, and Chen]{zheng2024genad}
Wenzhao Zheng, Ruiqi Song, Xianda Guo, and Long Chen.
\newblock Genad: Generative end-to-end autonomous driving.
\newblock In \emph{Proc. Eur. Conf. Comput. Vis. (ECCV)}, 2024.

\bibitem[Zhou et~al.(2024)Zhou, Lin, Shan, Wang, Sun, and Yang]{zhou2023drivinggaussian}
Xiaoyu Zhou, Zhiwei Lin, Xiaojun Shan, Yongtao Wang, Deqing Sun, and Ming-Hsuan Yang.
\newblock Drivinggaussian: Composite gaussian splatting for surrounding dynamic autonomous driving scenes.
\newblock In \emph{Proc. IEEE Conf. Comput. Vis. Pattern Recognit. (CVPR)}, 2024.

\end{thebibliography}
}
\clearpage
\setcounter{page}{1}

\twocolumn[{%
\renewcommand\twocolumn[1][]{#1}%
\maketitlesupplementary
\includegraphics[width=0.9\linewidth]{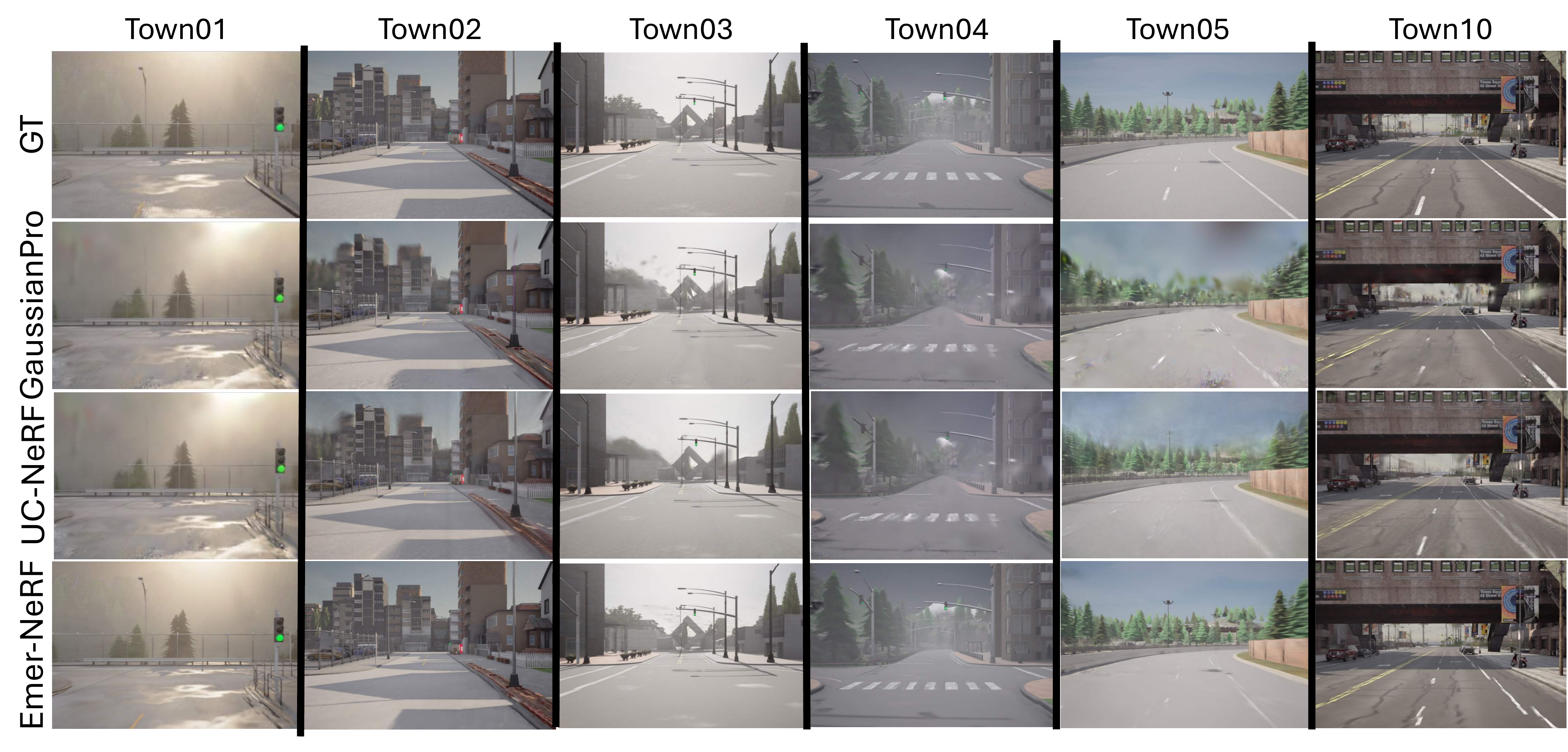}
\captionof{figure}{The ground truth and rendered results, obtained from various benchmarking methods, are compared across different scenes with a 1-meter offset.}
\label{fig:5}
\includegraphics[width=0.9\linewidth]{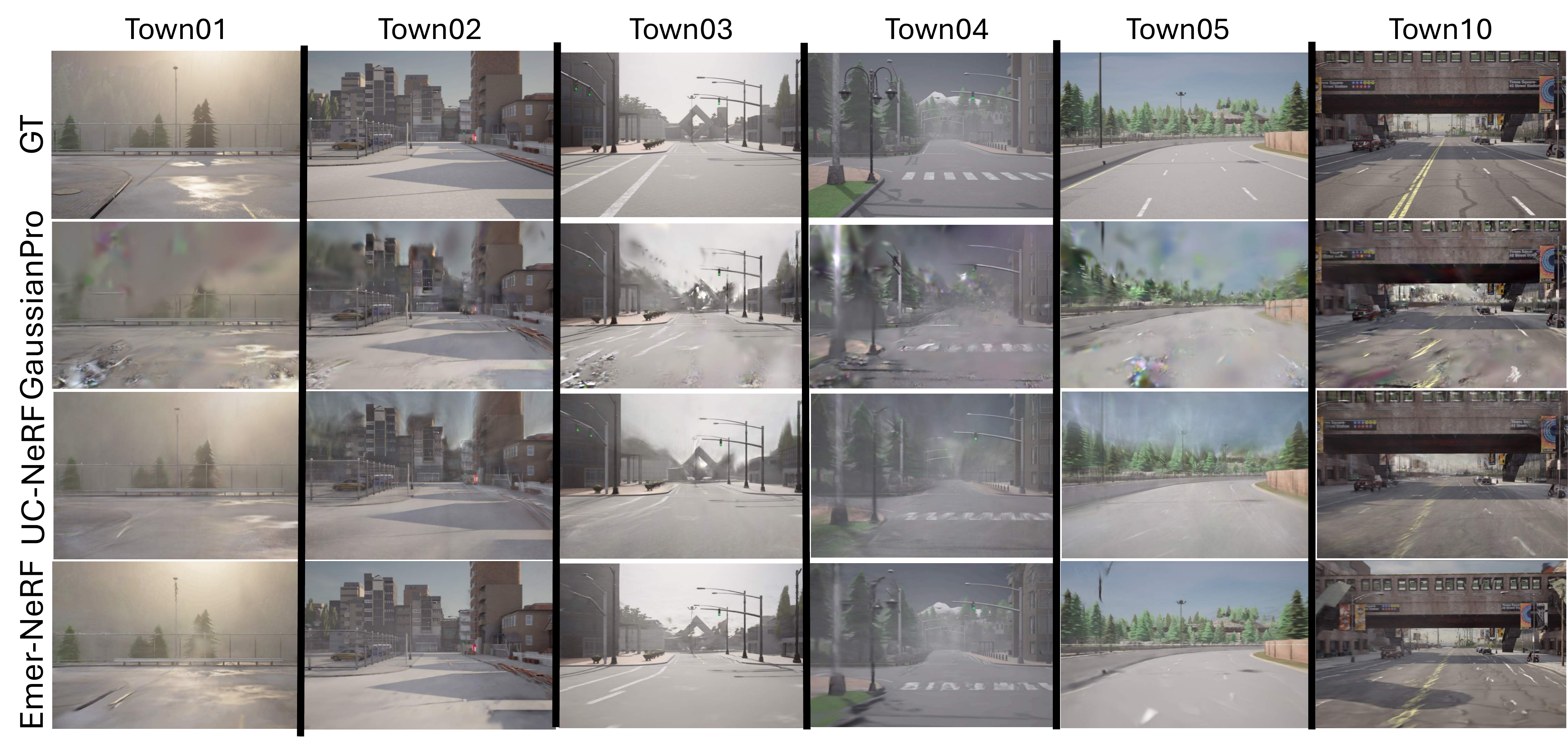}
\captionof{figure}{The ground truth and rendered results, obtained from various benchmarking methods, are compared across different scenes with a 4-meter offset.}
\label{fig:6}
}]





In this supplementary, we present the evaluation results of novel view synthesis using both front-only and multi-camera settings. For the front-only camera setup, we detail the results with various offsets (e.g., 0m, 1m, 2m, 4m), as illustrated in Tables \ref{tab:cam1-0m}, \ref{tab:cam1-1m}, \ref{tab:cam1-2m}, and \ref{tab:cam1-4m}. Additionally, we provide comprehensive experimental results for the multi-camera configuration, which are shown in Table \ref{tab:main_3_cam}. The novel view synthesis results for this setting, also with different offsets (e.g., 0m, 1m, 2m, 4m), are detailed in Tables \ref{tab:cam3-0m}, \ref{tab:cam3-1m}, \ref{tab:cam3-2m}, and \ref{tab:cam3-4m}.

\begin{figure*}[!t]
    \centering
    \includegraphics[width=0.75\linewidth]{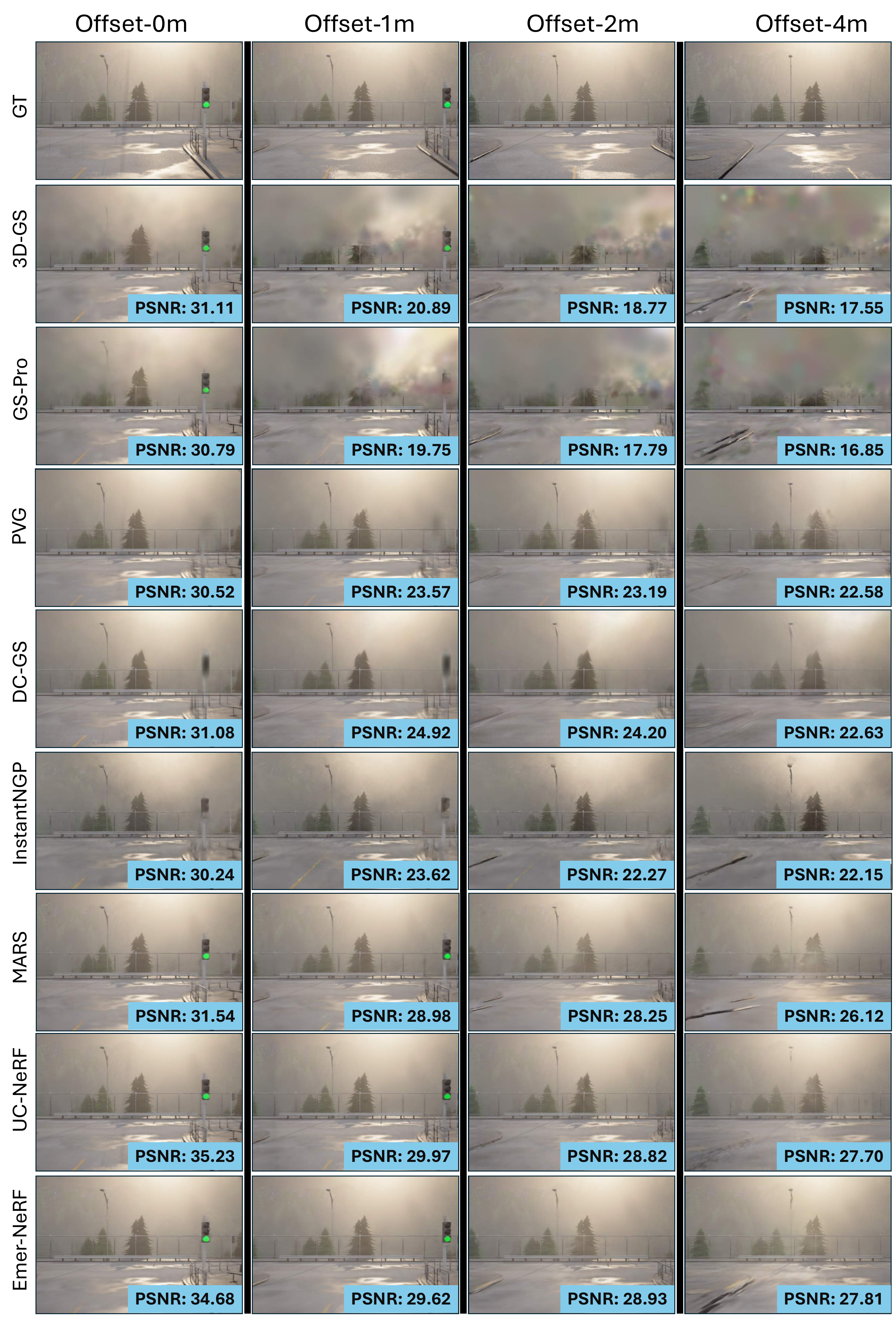}
    \caption{The ground truth and rendered results by different benchmarking methods from a sequence under different meters of offsetting.
}
    \label{fig:additional}
\end{figure*}

We also present additional results of various methods across different offsets for the Town-01 scene, as illustrated in Fig. \ref{fig:additional}. Furthermore, we showcase the visualization results for each scene using different methods. Specifically, Fig. \ref{fig:5} displays the results with a 1-meter offset, while Fig. \ref{fig:6} presents the results with a 4-meter offset.

\setlength{\tabcolsep}{3pt}
\begin{table*}[!b]
\centering
\footnotesize
\caption{Results on our proposed benchmark with the 0 meter offset with 1 camera.}
\label{tab:cam1-0m}
\begin{tabular}{lccccccccc}
\toprule 
&\multicolumn{3}{c}{Scene001} & \multicolumn{3}{c}{Scene002} & \multicolumn{3}{c}{Scene003} \\
& $\uparrow$PSNR & $\uparrow$SSIM & $\downarrow$LPIPS & $\uparrow$PSNR & $\uparrow$SSIM & $\downarrow$LPIPS & $\uparrow$PSNR & $\uparrow$SSIM & $\downarrow$LPIPS \\ \hline

\multicolumn{4}{l}{\small\textit{- NeRF-based}}\\
\hspace{1em}Instant-NGP~\cite{muller2022instant}\xspace 
& 31.64 & 0.919 & 0.214
& 30.95 & 0.881 & 0.265 
& 32.13 & 0.928 & 0.203 \\
\hspace{1em}UC-NeRF~\cite{cheng2023uc}\xspace 
& 39.35 & 0.958 & 0.315
& 35.19 & 0.941 & 0.288
& 34.53 & 0.939 & 0.303\\
\hspace{1em}MARS~\cite{wu2023mars}\xspace 
& 32.70 & 0.896 & 0.140
& 31.87 & 0.921 & 0.100
& 31.76 & 0.894 & 0.125\\
\hspace{1em}NeRFacto~\cite{tancik2023nerfstudio}\xspace 
& 28.74 & 0.905 & 0.253 
& 28.40 & 0.901 & 0.224 
& 31.25 & 0.924 & 0.190 \\
\hspace{1em}EmerNeRF~\cite{yang2023emernerf}\xspace 
& 33.43 & 0.932 & 0.111
& 31.72 & 0.914 & 0.106
& 32.33 & 0.922 & 0.102 \\
\hline
\multicolumn{4}{l}{\small\textit{- Gaussian-based}} \\
\hspace{1em}3D-GS~\cite{kerbl3Dgaussians}\xspace 
& 32.12 & 0.937 & 0.326 
& 29.09 & 0.908 & 0.295 
& 29.62 & 0.916 & 0.248 \\
\hspace{1em}PVG~\cite{chen2023periodic}\xspace
& 39.16 & 0.971 & 0.212
& 38.81 & 0.973 & 0.12
& 38.08 & 0.968 & 0.171\\
\hspace{1em}GaussianPro~\cite{cheng2024gaussianpro}\xspace 
& 35.51 & 0.953 & 0.266 
& 30.56 & 0.915 & 0.251 
& 29.39 & 0.910 & 0.242 \\

\hspace{1em}DC-Gaussian~\cite{wang2024dc}\xspace 
& 35.51 & 0.953 & 0.266
& 30.56 & 0.915 & 0.251
& 29.39 & 0.91 & 0.242\\
\hline

\toprule
&\multicolumn{3}{c}{Scene004} & \multicolumn{3}{c}{Scene005} & \multicolumn{3}{c}{Scene006} \\
& $\uparrow$PSNR & $\uparrow$SSIM & $\downarrow$LPIPS & $\uparrow$PSNR & $\uparrow$SSIM & $\downarrow$LPIPS & $\uparrow$PSNR & $\uparrow$SSIM & $\downarrow$LPIPS \\ \hline

\multicolumn{4}{l}{\small\textit{- NeRF-based}}\\
\hspace{1em}Instant-NGP~\cite{muller2022instant}\xspace 
& 29.58 & 0.921 & 0.221 
& 27.37 & 0.892 & 0.241 
& 26.86 & 0.820 & 0.372 \\
\hspace{1em}UC-NeRF~\cite{cheng2023uc}\xspace 
& 39.03 & 0.957 & 0.300
& 33.60 & 0.923 & 0.321
& 34.00 & 0.900 & 0.340\\
\hspace{1em}MARS~\cite{wu2023mars}\xspace 
& 36.68 & 0.957 & 0.084
& 26.31 & 0.857 & 0.188
& 30.44 & 0.875 & 0.151\\
\hspace{1em}NeRFacto~\cite{tancik2023nerfstudio}\xspace 
& 26.09 & 0.899 & 0.265 
& 23.64 & 0.871 & 0.261 
& 26.23 & 0.826 & 0.316 \\
\hspace{1em}EmerNeRF~\cite{yang2023emernerf}\xspace 
& 36.42 & 0.944 & 0.094
& 26.63 & 0.873 & 0.175
& 30.03 & 0.854 & 0.167 \\
\hline

\multicolumn{4}{l}{\small\textit{- Gaussian-based}} \\
\hspace{1em}3D-GS~\cite{kerbl3Dgaussians}\xspace  
& 33.60 & 0.946 & 0.280 
& 25.95 & 0.872 & 0.380 
& 28.45 & 0.841 & 0.390 \\
\hspace{1em}PVG~\cite{chen2023periodic}\xspace
& 39.58 & 0.971 & 0.194
& 34.91 & 0.954 & 0.194
& 36.16 & 0.921 & 0.244\\
\hspace{1em}GaussianPro~\cite{cheng2024gaussianpro}\xspace 
& 33.69 & 0.943 & 0.273 
& 30.54 & 0.916 & 0.249 
& 30.03 & 0.879 & 0.301 \\
\hspace{1em}DC-Gaussian~\cite{wang2024dc}\xspace 
& 33.69 & 0.943 & 0.273
& 26.17 & 0.916 & 0.249
& 30.03 & 0.879 & 0.301\\
\bottomrule
\end{tabular}
\end{table*}

\setlength{\tabcolsep}{3pt}
\begin{table*}[!b]
\vspace{-10pt}
\centering
\footnotesize
\caption{Results on our proposed benchmark with the 1-meter offset with 1 camera.}
\label{tab:cam1-1m}
\begin{tabular}{lccccccccc}
\toprule 
&\multicolumn{3}{c}{Scene001} & \multicolumn{3}{c}{Scene002} & \multicolumn{3}{c}{Scene003} \\
& $\uparrow$PSNR & $\uparrow$SSIM & $\downarrow$LPIPS & $\uparrow$PSNR & $\uparrow$SSIM & $\downarrow$LPIPS & $\uparrow$PSNR & $\uparrow$SSIM & $\downarrow$LPIPS \\ \hline

\multicolumn{4}{l}{\small\textit{- NeRF-based}}\\
\hspace{1em}Instant-NGP~\cite{muller2022instant}\xspace 
& 24.19 & 0.868 & 0.350
& 20.25 & 0.766 & 0.390
& 23.92 & 0.841 & 0.288 \\
\hspace{1em}UC-NeRF~\cite{cheng2023uc}\xspace 
& 29.60 & 0.925 & 0.355
& 28.68 & 0.894 & 0.338
& 30.11 & 0.911 & 0.336\\

\hspace{1em}MARS~\cite{wu2023mars}\xspace 
& 28.27 & 0.886 & 0.145
& 27.95 & 0.883 & 0.138
& 29.04 & 0.899 & 0.140\\
\hspace{1em}NeRFacto~\cite{tancik2023nerfstudio}\xspace 
& 24.48 & 0.884 & 0.300 
& 24.40 & 0.842 & 0.279
& 23.55 & 0.835 & 0.278 \\

\hspace{1em}EmerNeRF~\cite{yang2023emernerf}\xspace 
& 28.81 & 0.913 & 0.123
& 27.82 & 0.876 & 0.143
& 29.39 & 0.901 & 0.126 \\
\hline
\multicolumn{4}{l}{\small\textit{- Gaussian-based}} \\
\hspace{1em}3D-GS~\cite{kerbl3Dgaussians}\xspace 
& 24.07 & 0.909 & 0.321
& 19.98 & 0.799 & 0.372
& 21.22 & 0.811 & 0.334 \\
\hspace{1em}PVG~\cite{chen2023periodic}\xspace
& 25.62 & 0.917 & 0.282
& 27.25 & 0.912 & 0.267
& 28.15 & 0.907 & 0.253\\
\hspace{1em}GaussianPro~\cite{cheng2024gaussianpro}\xspace 
& 23.97 & 0.902 & 0.329 
& 20.71 & 0.811 & 0.360 
& 21.92 & 0.840 & 0.322 \\
\hspace{1em}DC-Gaussian~\cite{wang2024dc}\xspace 
& 27.34 & 0.923 & 0.272
& 24.87 & 0.873 & 0.303
& 26.35 & 0.891 & 0.275\\
\hline

\toprule
&\multicolumn{3}{c}{Scene004} & \multicolumn{3}{c}{Scene005} & \multicolumn{3}{c}{Scene006} \\
& $\uparrow$PSNR & $\uparrow$SSIM & $\downarrow$LPIPS & $\uparrow$PSNR & $\uparrow$SSIM & $\downarrow$LPIPS & $\uparrow$PSNR & $\uparrow$SSIM & $\downarrow$LPIPS \\ \hline

\multicolumn{4}{l}{\small\textit{- NeRF-based}}\\
\hspace{1em}Instant-NGP~\cite{muller2022instant}\xspace 
& 26.21 & 0.869 & 0.277
& 22.18 & 0.800 & 0.338
& 23.88 & 0.739 & 0.434 \\
\hspace{1em}UC-NeRF~\cite{cheng2023uc}\xspace 
& 34.10 & 0.938 & 0.338
& 29.42 & 0.890 & 0.353
& 28.51 & 0.820 & 0.410\\
\hspace{1em}MARS~\cite{wu2023mars}\xspace 
& 34.28 & 0.937 & 0.097
& 24.42 & 0.831 & 0.210
& 27.25 & 0.812 & 0.199\\
\hspace{1em}NeRFacto~\cite{tancik2023nerfstudio}\xspace 
& 24.20 & 0.858 & 0.311
& 21.00 & 0.794 & 0.336 
& 23.30 & 0.732 & 0.380 \\
\hspace{1em}EmerNeRF~\cite{yang2023emernerf}\xspace 
& 34.15 & 0.930 & 0.102
& 24.83 & 0.851 & 0.194
& 26.96 & 0.797 & 0.211 \\

\hline

\multicolumn{4}{l}{\small\textit{- Gaussian-based}} \\
\hspace{1em}3D-GS~\cite{kerbl3Dgaussians}\xspace  
& 25.62 & 0.905 & 0.343
& 21.52 & 0.840 & 0.346
& 21.75 & 0.843 & 0.342\\
\hspace{1em}PVG~\cite{chen2023periodic}\xspace
& 29.19 & 0.926 & 0.286
& 25.33 & 0.864 & 0.293
& 25.51 & 0.764 & 0.397\\
\hspace{1em}GaussianPro~\cite{cheng2024gaussianpro}\xspace 
& 25.56 & 0.907 & 0.341 
& 21.52 & 0.840 & 0.346 
& 21.92 & 0.841 & 0.335 \\
\hspace{1em}DC-Gaussian~\cite{wang2024dc}\xspace 
& 30.67 & 0.936 & 0.266
& 26.17 & 0.883 & 0.297
& 25.51 & 0.796 & 0.376\\
\bottomrule
\end{tabular}
\end{table*}

\setlength{\tabcolsep}{3pt}
\begin{table*}[!b]
\centering
\footnotesize
\caption{Results on our proposed benchmark with the 2-meter offset with 1 camera.}
\label{tab:cam1-2m}
\begin{tabular}{lccccccccc}
\toprule 
&\multicolumn{3}{c}{Scene001} & \multicolumn{3}{c}{Scene002} & \multicolumn{3}{c}{Scene003} \\
& $\uparrow$PSNR & $\uparrow$SSIM & $\downarrow$LPIPS & $\uparrow$PSNR & $\uparrow$SSIM & $\downarrow$LPIPS & $\uparrow$PSNR & $\uparrow$SSIM & $\downarrow$LPIPS \\ \hline

\multicolumn{4}{l}{\small\textit{- NeRF-based}}\\
\hspace{1em}Instant-NGP~\cite{muller2022instant}\xspace 
& 24.21 & 0.861 & 0.377
& 20.14 & 0.767 & 0.405
& 21.61 & 0.802 & 0.344 \\
\hspace{1em}UC-NeRF~\cite{cheng2023uc}\xspace 
& 28.45 & 0.917 & 0.373
& 27.53 & 0.895 & 0.347
& 29.04 & 0.908 & 0.343\\
\hspace{1em}MARS~\cite{wu2023mars}\xspace 
& 26.58 & 0.875 & 0.165
& 27.29 & 0.886 & 0.132
& 27.37 & 0.886 & 0.151\\

\hspace{1em}NeRFacto~\cite{tancik2023nerfstudio}\xspace 
& 23.73 & 0.869 & 0.334
& 22.95 & 0.818 & 0.312
& 21.29 & 0.795 & 0.333 \\

\hspace{1em}EmerNeRF~\cite{yang2023emernerf}\xspace 
& 27.12 & 0.902 & 0.143
& 27.16 & 0.879 & 0.137
& 27.72 & 0.888 & 0.137 \\
\hline
\multicolumn{4}{l}{\small\textit{- Gaussian-based}} \\
\hspace{1em}3D-GS~\cite{kerbl3Dgaussians}\xspace 
& 22.00 & 0.881 & 0.369
& 19.28 & 0.796 & 0.394
& 21.91 & 0.840 & 0.322\\
\hspace{1em}PVG~\cite{chen2023periodic}\xspace
& 24.58 & 0.910 & 0.306
& 26.13 & 0.885 & 0.293
& 26.58 & 0.894 & 0.272\\
\hspace{1em}GaussianPro~\cite{cheng2024gaussianpro}\xspace 
& 21.89 & 0.879 & 0.368
& 19.28 & 0.796 & 0.394
& 21.91 & 0.84 & 0.322\\
\hspace{1em}DC-Gaussian~\cite{wang2024dc}\xspace 
& 25.89 & 0.912 & 0.294 
& 23.28 & 0.866 & 0.314 
& 24.61 & 0.876 & 0.299 \\
\hline

\toprule
&\multicolumn{3}{c}{Scene004} & \multicolumn{3}{c}{Scene005} & \multicolumn{3}{c}{Scene006} \\
& $\uparrow$PSNR & $\uparrow$SSIM & $\downarrow$LPIPS & $\uparrow$PSNR & $\uparrow$SSIM & $\downarrow$LPIPS & $\uparrow$PSNR & $\uparrow$SSIM & $\downarrow$LPIPS \\ \hline

\multicolumn{4}{l}{\small\textit{- NeRF-based}}\\
\hspace{1em}Instant-NGP~\cite{muller2022instant}\xspace 
& 24.71 & 0.841 & 0.327 
& 21.03 & 0.776 & 0.387
& 22.5 & 0.694 & 0.474 \\
\hspace{1em}UC-NeRF~\cite{cheng2023uc}\xspace 
& 31.31 & 0.925 & 0.353
& 28.46 & 0.882 & 0.363
& 26.95 & 0.907 & 0.419\\
\hspace{1em}MARS~\cite{wu2023mars}\xspace 
& 31.57 & 0.924 & 0.120
& 24.19 & 0.819 & 0.222
& 26.12 & 0.795 & 0.214\\
\hspace{1em}NeRFacto~\cite{tancik2023nerfstudio}\xspace 
& 24.54 & 0.873 & 0.301
& 20.25 & 0.772 & 0.381
& 21.97 & 0.685 & 0.419\\
\hspace{1em}EmerNeRF~\cite{yang2023emernerf}\xspace 
& 31.44 & 0.917 & 0.125
& 24.60 & 0.839 & 0.206
& 25.83 & 0.780 & 0.226 \\
\hline

\multicolumn{4}{l}{\small\textit{- Gaussian-based}} \\
\hspace{1em}3D-GS~\cite{kerbl3Dgaussians}\xspace  
& 23.35 & 0.878 & 0.384
& 20.52 & 0.817 & 0.381
& 20.53 & 0.806 & 0.458\\
\hspace{1em}PVG~\cite{chen2023periodic}\xspace
& 28.27 & 0.914 & 0.312
& 24.41 & 0.853 & 0.311
& 24.13 & 0.744 & 0.411\\
\hspace{1em}GaussianPro~\cite{cheng2024gaussianpro}\xspace 
& 23.54 & 0.881 & 0.381
& 20.52 & 0.817 & 0.381
& 20.45 & 0.706 & 0.457\\
\hspace{1em}DC-Gaussian~\cite{wang2024dc}\xspace 
& 28.27 & 0.920 & 0.294 
& 25.21 & 0.879 & 0.321 
& 24.19 & 0.773 & 0.392 \\
\bottomrule
\end{tabular}
\end{table*}

\setlength{\tabcolsep}{3pt}
\begin{table*}[!b]
\centering
\footnotesize
\caption{Results on our proposed benchmark with the 4-meter offset with 1 camera.}
\label{tab:cam1-4m}
\begin{tabular}{lccccccccc}
\toprule 
&\multicolumn{3}{c}{Scene001} & \multicolumn{3}{c}{Scene002} & \multicolumn{3}{c}{Scene003} \\
& $\uparrow$PSNR & $\uparrow$SSIM & $\downarrow$LPIPS & $\uparrow$PSNR & $\uparrow$SSIM & $\downarrow$LPIPS & $\uparrow$PSNR & $\uparrow$SSIM & $\downarrow$LPIPS \\ \hline

\multicolumn{4}{l}{\small\textit{- NeRF-based}}\\
\hspace{1em}Instant-NGP~\cite{muller2022instant}\xspace 
& 23.20 & 0.854 & 0.401
& 19.44 & 0.754 & 0.477 
& 19.49 & 0.764 & 0.413 \\

\hspace{1em}UC-NeRF~\cite{cheng2023uc}\xspace 
& 26.12 & 0.903 & 0.410
& 24.34 & 0.872 & 0.400
& 27.23 & 0.888 & 0.373\\
\hspace{1em}MARS~\cite{wu2023mars}\xspace 
& 23.45 & 0.854 & 0.202
& 24.41 & 0.848 & 0.177
& 25.00 & 0.843 & 0.196\\
\hspace{1em}NeRFacto~\cite{tancik2023nerfstudio}\xspace 
& 21.70 & 0.847 & 0.366 
& 20.14 & 0.767 & 0.405
& 19.15 & 0.756 & 0.402 \\
\hspace{1em}EmerNeRF~\cite{yang2023emernerf}\xspace 
& 23.89 & 0.876 & 0.184
& 24.31 & 0.843 & 0.181
& 25.39 & 0.862 & 0.180 \\
\hline
\multicolumn{4}{l}{\small\textit{- Gaussian-based}} \\
\hspace{1em}3D-GS~\cite{kerbl3Dgaussians}\xspace 
& 19.57 & 0.842 & 0.424
& 17.42 & 0.755 & 0.449
& 18.01 & 0.773 & 0.414\\
\hspace{1em}PVG~\cite{chen2023periodic}\xspace
& 22.59 & 0.893 & 0.336
& 22.74 & 0.862 & 0.327
& 23.65 & 0.866 & 0.310\\
\hspace{1em}GaussianPro~\cite{cheng2024gaussianpro}\xspace 
& 19.50 & 0.840 & 0.420 
& 17.39 & 0.753 & 0.447 
& 17.96 & 0.767 & 0.411 \\
\hspace{1em}DC-Gaussian~\cite{wang2024dc}\xspace 
& 23.56 & 0.897 & 0.329
& 22.35 & 0.835 & 0.362 
& 22.06 & 0.854 & 0.338 \\
\hline

\toprule
&\multicolumn{3}{c}{Scene004} & \multicolumn{3}{c}{Scene005} & \multicolumn{3}{c}{Scene006} \\
& $\uparrow$PSNR & $\uparrow$SSIM & $\downarrow$LPIPS & $\uparrow$PSNR & $\uparrow$SSIM & $\downarrow$LPIPS & $\uparrow$PSNR & $\uparrow$SSIM & $\downarrow$LPIPS \\ \hline

\multicolumn{4}{l}{\small\textit{- NeRF-based}}\\
\hspace{1em}Instant-NGP~\cite{muller2022instant}\xspace 
& 22.59 & 0.817 & 0.409 
& 19.95 & 0.764 & 0.422
& 21.02 & 0.652 & 0.533 \\
\hspace{1em}UC-NeRF~\cite{cheng2023uc}\xspace 
& 27.96 & 0.890 & 0.409
& 26.27 & 0.852 & 0.397
& 26.07 & 0.755 & 0.473\\
\hspace{1em}MARS~\cite{wu2023mars}\xspace 

& 27.39 & 0.888 & 0.175
& 23.37 & 0.813 & 0.239
& 24.36 & 0.749 & 0.258\\
\hspace{1em}NeRFacto~\cite{tancik2023nerfstudio}\xspace 
& 20.90 & 0.815 & 0.419
& 19.49 & 0.760 & 0.414 
& 20.39 & 0.637 & 0.489 \\
\hspace{1em}EmerNeRF~\cite{yang2023emernerf}\xspace 
& 27.35 & 0.886 & 0.177
& 23.66 & 0.814 & 0.227
& 24.17 & 0.739 & 0.266 \\
\hline

\multicolumn{4}{l}{\small\textit{- Gaussian-based}} \\
\hspace{1em}3D-GS~\cite{kerbl3Dgaussians}\xspace  
& 20.39 & 0.835 & 0.451
& 19.01 & 0.795 & 0.426
& 18.42 & 0.648 & 0.515\\
\hspace{1em}PVG~\cite{chen2023periodic}\xspace
& 25.11 & 0.891 & 0.359
& 22.90 & 0.831 & 0.338
& 22.00 & 0.700 & 0.447\\
\hspace{1em}GaussianPro~\cite{cheng2024gaussianpro}\xspace 
& 20.39 & 0.835 & 0.451
& 18.91 & 0.783 & 0.43
& 18.33 & 0.647 & 0.511\\
\hspace{1em}DC-Gaussian~\cite{wang2024dc}\xspace 
& 24.26 & 0.893 & 0.345 
& 23.18 & 0.850 & 0.356 
& 21.99 & 0.737 & 0.427 \\
\bottomrule
\end{tabular}
\end{table*}

\setlength{\tabcolsep}{3pt}
\begin{table*}[!b]
\centering
\footnotesize
\caption{Results on our proposed benchmark with the 0-meter offset with 3 cameras.}
\label{tab:cam3-0m}
\begin{tabular}{lccccccccc}
\toprule 
&\multicolumn{3}{c}{Scene001} & \multicolumn{3}{c}{Scene002} & \multicolumn{3}{c}{Scene003} \\
& $\uparrow$PSNR & $\uparrow$SSIM & $\downarrow$LPIPS & $\uparrow$PSNR & $\uparrow$SSIM & $\downarrow$LPIPS & $\uparrow$PSNR & $\uparrow$SSIM & $\downarrow$LPIPS \\ \hline

\multicolumn{4}{l}{\small\textit{- NeRF-based}}\\
\hspace{1em}Instant-NGP~\cite{muller2022instant}\xspace 
& 29.56 & 0.906 & 0.250
& 30.95 & 0.907 & 0.236
& 31.13 & 0.911 & 0.217 \\
\hspace{1em}UC-NeRF~\cite{cheng2023uc}\xspace 
& 35.43 & 0.936 & 0.375
& 31.40 & 0.904 & 0.365
& 31.95 & 0.921 & 0.344\\
\hspace{1em}MARS~\cite{wu2023mars}\xspace 
& 32.52 & 0.890 & 0.154
& 30.99 & 0.903 & 0.133
& 31.99 & 0.889 & 0.136 \\
\hspace{1em}NeRFacto~\cite{tancik2023nerfstudio}\xspace 
& 30.75 & 0.902 & 0.198 
& 29.64 & 0.894 & 0.238 
& 30.83 & 0.907 & 0.214 \\
\hspace{1em}EmerNeRF~\cite{yang2023emernerf}\xspace 
& 33.25 & 0.926 & 0.125
& 30.84 & 0.896 & 0.139
& 32.56 & 0.917 & 0.113 \\

\hline

\multicolumn{4}{l}{\small\textit{- Gaussian-based}} \\
\hspace{1em}3D-GS~\cite{kerbl3Dgaussians}\xspace 
& 31.35 & 0.935 & 0.310
& 27.09 & 0.883 & 0.315
& 27.23 & 0.892 & 0.278\\
\hspace{1em}PVG~\cite{chen2023periodic}\xspace
& 34.59 & 0.955 & 0.258
& 33.77 & 0.948 & 0.223
& 34.08 & 0.948 & 0.219\\
\hspace{1em}GaussianPro~\cite{cheng2024gaussianpro}\xspace 
& 31.07 & 0.934 & 0.211 
& 27.23 & 0.883 & 0.215 
& 26.97 & 0.888 & 0.283 \\
\hspace{1em}DC-Gaussian~\cite{wang2024dc}\xspace 
& 33.95 & 0.948 & 0.258
& 29.19 & 0.910 & 0.262
& 28.58 & 0.914 & 0.251\\
\hline

\toprule
&\multicolumn{3}{c}{Scene004} & \multicolumn{3}{c}{Scene005} & \multicolumn{3}{c}{Scene006} \\
& $\uparrow$PSNR & $\uparrow$SSIM & $\downarrow$LPIPS & $\uparrow$PSNR & $\uparrow$SSIM & $\downarrow$LPIPS & $\uparrow$PSNR & $\uparrow$SSIM & $\downarrow$LPIPS \\ \hline

\multicolumn{4}{l}{\small\textit{- NeRF-based}}\\
\hspace{1em}Instant-NGP~\cite{muller2022instant}\xspace 
& 29.73 & 0.899 & 0.222 
& 26.69 & 0.871 & 0.278 
& 27.39 & 0.836 & 0.366 \\
\hspace{1em}UC-NeRF~\cite{cheng2023uc}\xspace 
& 36.29 & 0.942 & 0.332
& 32.51 & 0.909 & 0.340
& 31.15 & 0.859 & 0.403\\
\hspace{1em}MARS~\cite{wu2023mars}\xspace 
& 35.72 & 0.942 & 0.111
& 28.28 & 0.856 & 0.170
& 28.73 & 0.843 & 0.203 \\
\hspace{1em}NeRFacto~\cite{tancik2023nerfstudio}\xspace 
& 29.65 & 0.887 & 0.268 
& 27.46 & 0.898 & 0.277 
& 28.00 & 0.853 & 0.281 \\
\hspace{1em}EmerNeRF~\cite{yang2023emernerf}\xspace 
& 35.46 & 0.929 & 0.121
& 28.60 & 0.872 & 0.157
& 28.32 & 0.822 & 0.219\\
\hline

\multicolumn{4}{l}{\small\textit{- Gaussian-based}} \\
\hspace{1em}3D-GS~\cite{kerbl3Dgaussians}\xspace  
& 30.62 & 0.924 & 0.320
& 31.25 & 0.930 & 0.321
& 30.91 & 0.922 & 0.326\\
\hspace{1em}PVG~\cite{chen2023periodic}\xspace
& 35.41 & 0.956 & 0.243
& 30.79 & 0.916 & 0.263
& 31.36 & 0.874 & 0.329\\
\hspace{1em}GaussianPro~\cite{cheng2024gaussianpro}\xspace 
& 30.45 & 0.923 & 0.322 
& 25.69 & 0.871 & 0.328 
& 27.39 & 0.836 & 0.366 \\
\hspace{1em}DC-Gaussian~\cite{wang2024dc}\xspace 
& 32.74 & 0.947 & 0.251
& 27.91 & 0.898 & 0.275
& 29.01 & 0.857 & 0.327\\
\bottomrule
\end{tabular}
\end{table*}

\setlength{\tabcolsep}{3pt}
\begin{table*}[!b]
\centering
\footnotesize
\caption{Results on our proposed benchmark with the 1-meter offset with 3 cameras.}
\label{tab:cam3-1m}
\begin{tabular}{lccccccccc}
\toprule 
&\multicolumn{3}{c}{Scene001} & \multicolumn{3}{c}{Scene002} & \multicolumn{3}{c}{Scene003} \\
& $\uparrow$PSNR & $\uparrow$SSIM & $\downarrow$LPIPS & $\uparrow$PSNR & $\uparrow$SSIM & $\downarrow$LPIPS & $\uparrow$PSNR & $\uparrow$SSIM & $\downarrow$LPIPS \\ \hline

\multicolumn{4}{l}{\small\textit{- NeRF-based}}\\
\hspace{1em}Instant-NGP~\cite{muller2022instant}\xspace 
& 22.51 & 0.881 & 0.361
& 20.41 & 0.794 & 0.358
& 23.62 & 0.892 & 0.263\\
\hspace{1em}UC-NeRF~\cite{cheng2023uc}\xspace 
& 29.60 & 0.925 & 0.355
& 28.68 & 0.894 & 0.338
& 30.11 & 0.911 & 0.336\\
\hspace{1em}MARS~\cite{wu2023mars}\xspace 
& 28.58 & 0.888 & 0.149
& 28.81 & 0.884 & 0.144
& 30.10 & 0.907 & 0.129\\
\hspace{1em}NeRFacto~\cite{tancik2023nerfstudio}\xspace 
& 23.60 & 0.891 & 0.311
& 22.70 & 0.862 & 0.272
& 21.98 & 0.826 & 0.286\\
\hspace{1em}EmerNeRF~\cite{yang2023emernerf}\xspace 
& 29.12 & 0.915 & 0.127
& 28.68 & 0.877 & 0.149
& 30.448 & 0.909 & 0.115 \\
\hline
\multicolumn{4}{l}{\small\textit{- Gaussian-based}} \\
\hspace{1em}3D-GS~\cite{kerbl3Dgaussians}\xspace 
& 23.90 & 0.905 & 0.340
& 20.67 & 0.818 & 0.382
& 22.33 & 0.846 & 0.331\\
\hspace{1em}PVG~\cite{chen2023periodic}\xspace
& 25.48 & 0.92 & 0.290
& 25.82 & 0.884 & 0.295
& 28.11 & 0.907 & 0.270\\
\hspace{1em}GaussianPro~\cite{cheng2024gaussianpro}\xspace 
& 23.37 & 0.902 & 0.349 
& 20.81 & 0.817 & 0.353 
& 22.19 & 0.0.740 & 0.424 \\
\hspace{1em}DC-Gaussian~\cite{wang2024dc}\xspace 
& 28.01 & 0.928 & 0.288
& 25.08 & 0.869 & 0.315
& 25.92 & 0.891 & 0.289\\
\hline

\toprule
&\multicolumn{3}{c}{Scene004} & \multicolumn{3}{c}{Scene005} & \multicolumn{3}{c}{Scene006} \\
& $\uparrow$PSNR & $\uparrow$SSIM & $\downarrow$LPIPS & $\uparrow$PSNR & $\uparrow$SSIM & $\downarrow$LPIPS & $\uparrow$PSNR & $\uparrow$SSIM & $\downarrow$LPIPS \\ \hline

\multicolumn{4}{l}{\small\textit{- NeRF-based}}\\
\hspace{1em}Instant-NGP~\cite{muller2022instant}\xspace 
& 26.60 & 0.897 & 0.287
& 23.15 & 0.854 & 0.321
& 24.81 & 0.763 & 0.472\\
\hspace{1em}UC-NeRF~\cite{cheng2023uc}\xspace 
& 34.10 & 0.938 & 0.338
& 29.42 & 0.890 & 0.353
& 28.51 & 0.820 & 0.410\\
\hspace{1em}MARS~\cite{wu2023mars}\xspace 
& 33.96 & 0.930 & 0.110
& 26.42 & 0.832 & 0.190
& 27.84 & 0.807 & 0.220\\
\hspace{1em}NeRFacto~\cite{tancik2023nerfstudio}\xspace 
& 24.08 & 0.869 & 0.300
& 20.50 & 0.793 & 0.351
& 24.58 & 0.861 & 0.319\\
\hspace{1em}EmerNeRF~\cite{yang2023emernerf}\xspace 
& 33.83 & 0.923 & 0.115
& 26.83 & 0.852 & 0.174
& 27.55 & 0.792 & 0.232 \\
\hline

\multicolumn{4}{l}{\small\textit{- Gaussian-based}} \\
\hspace{1em}3D-GS~\cite{kerbl3Dgaussians}\xspace  
& 23.19 & 0.896 & 0.377
& 19.66 & 0.827 & 0.385
& 22.72 & 0.757 & 0.336\\
\hspace{1em}PVG~\cite{chen2023periodic}\xspace
& 29.08 & 0.926 & 0.305
& 25.54 & 0.864 & 0.319
& 25.69 & 0.769 & 0.428\\
\hspace{1em}GaussianPro~\cite{cheng2024gaussianpro}\xspace 
& 23.45 & 0.896 & 0.379
& 19.50 & 0.819 & 0.389 
& 22.69 & 0.757 & 0.443 \\
\hspace{1em}DC-Gaussian~\cite{wang2024dc}\xspace 
& 30.27 & 0.935 & 0.289
& 25.85 & 0.878 & 0.306
& 25.33 & 0.794 & 0.402\\
\bottomrule
\end{tabular}
\end{table*}

\setlength{\tabcolsep}{3pt}
\begin{table*}[!b]
\centering
\footnotesize
\caption{Results on our proposed benchmark with the 2-meter offset with 3 cameras.}
\label{tab:cam3-2m}
\begin{tabular}{lccccccccc}
\toprule 
&\multicolumn{3}{c}{Scene001} & \multicolumn{3}{c}{Scene002} & \multicolumn{3}{c}{Scene003} \\
& $\uparrow$PSNR & $\uparrow$SSIM & $\downarrow$LPIPS & $\uparrow$PSNR & $\uparrow$SSIM & $\downarrow$LPIPS & $\uparrow$PSNR & $\uparrow$SSIM & $\downarrow$LPIPS \\ \hline

\multicolumn{4}{l}{\small\textit{- NeRF-based}}\\
\hspace{1em}Instant-NGP~\cite{muller2022instant}\xspace 
& 22.51 & 0.881 & 0.361
& 20.41 & 0.794 & 0.358
& 23.62 & 0.892 & 0.263\\

\hspace{1em}UC-NeRF~\cite{cheng2023uc}\xspace 
& 29.60 & 0.925 & 0.355
& 28.68 & 0.894 & 0.338
& 30.11 & 0.911 & 0.336\\
\hspace{1em}MARS~\cite{wu2023mars}\xspace 
& 28.09 & 0.882 & 0.161
& 28.69 & 0.892 & 0.132
& 29.49 & 0.904 & 0.131\\

\hspace{1em}NeRFacto~\cite{tancik2023nerfstudio}\xspace 
& 23.57 & 0.842 & 0.323
& 19.97 & 0.769 & 0.393
& 23.86 & 0.728 & 0.408\\
\hspace{1em}EmerNeRF~\cite{yang2023emernerf}\xspace 
& 28.63 & 0.909 & 0.139
& 28.56 & 0.885 & 0.137
& 29.84 & 0.906 & 0.117 \\

\hline
\multicolumn{4}{l}{\small\textit{- Gaussian-based}} \\
\hspace{1em}3D-GS~\cite{kerbl3Dgaussians}\xspace 
& 22.43 & 0.891 & 0.369
& 22.21 & 0.812 & 0.397
& 21.15 & 0.825 & 0.355\\
\hspace{1em}PVG~\cite{chen2023periodic}\xspace
& 24.31 & 0.912 & 0.313
& 24.74 & 0.885 & 0.298
& 26.63 & 0.894 & 0.289\\
\hspace{1em}GaussianPro~\cite{cheng2024gaussianpro}\xspace 
& 22.39 & 0.891 & 0.368
& 20.06 & 0.812 & 0.396
& 20.76 & 0.819 & 0.362\\
\hspace{1em}DC-Gaussian~\cite{wang2024dc}\xspace 
& 26.98 & 0.921 & 0.303
& 24.02 & 0.871 & 0.315
& 24.51 & 0.878 & 0.306\\

\hline

\toprule
&\multicolumn{3}{c}{Scene004} & \multicolumn{3}{c}{Scene005} & \multicolumn{3}{c}{Scene006} \\
& $\uparrow$PSNR & $\uparrow$SSIM & $\downarrow$LPIPS & $\uparrow$PSNR & $\uparrow$SSIM & $\downarrow$LPIPS & $\uparrow$PSNR & $\uparrow$SSIM & $\downarrow$LPIPS \\ \hline

\multicolumn{4}{l}{\small\textit{- NeRF-based}}\\
\hspace{1em}Instant-NGP~\cite{muller2022instant}\xspace 
& 26.60 & 0.897 & 0.287
& 23.15 & 0.854 & 0.321
& 24.81 & 0.763 & 0.472\\
\hspace{1em}UC-NeRF~\cite{cheng2023uc}\xspace 
& 34.10 & 0.938 & 0.338
& 29.42 & 0.890 & 0.353
& 28.51 & 0.820 & 0.410\\
\hspace{1em}MARS~\cite{wu2023mars}\xspace 
& 31.95 & 0.919 & 0.131
& 25.77 & 0.821 & 0.205
& 27.22 & 0.798 & 0.228\\
\hspace{1em}NeRFacto~\cite{tancik2023nerfstudio}\xspace 
& 23.42 & 0.878 & 0.337
& 22.52 & 0.853 & 0.282
& 20.22 & 0.785 & 0.343\\
\hspace{1em}EmerNeRF~\cite{yang2023emernerf}\xspace 
& 31.82 & 0.912 & 0.136
& 26.18 & 0.841 & 0.189
& 26.93 & 0.783 & 0.240 \\
\hline

\multicolumn{4}{l}{\small\textit{- Gaussian-based}} \\
\hspace{1em}3D-GS~\cite{kerbl3Dgaussians}\xspace  
& 21.80 & 0.876 & 0.410
& 19.11 & 0.810 & 0.412
& 21.33 & 0.729 & 0.468\\
\hspace{1em}PVG~\cite{chen2023periodic}\xspace
& 28.10 & 0.917 & 0.325
& 24.65 & 0.855 & 0.333
& 24.58 & 0.756 & 0.435\\
\hspace{1em}GaussianPro~\cite{cheng2024gaussianpro}\xspace 
& 21.69 & 0.876 & 0.408
& 18.70 & 0.804 & 0.426
& 21.48 & 0.732 & 0.465\\
\hspace{1em}DC-Gaussian~\cite{wang2024dc}\xspace 
& 28.25 & 0.919 & 0.313
& 25.12 & 0.862 & 0.324
& 24.32 & 0.778 & 0.411\\
\bottomrule
\end{tabular}
\end{table*}

\setlength{\tabcolsep}{3pt}
\begin{table*}[!b]
\centering
\footnotesize
\caption{Results on our proposed benchmark with the 4-meter offset with 3 cameras.}
\label{tab:cam3-4m}
\begin{tabular}{lccccccccc}
\toprule 
&\multicolumn{3}{c}{Scene001} & \multicolumn{3}{c}{Scene002} & \multicolumn{3}{c}{Scene003} \\
& $\uparrow$PSNR & $\uparrow$SSIM & $\downarrow$LPIPS & $\uparrow$PSNR & $\uparrow$SSIM & $\downarrow$LPIPS & $\uparrow$PSNR & $\uparrow$SSIM & $\downarrow$LPIPS \\ \hline

\multicolumn{4}{l}{\small\textit{- NeRF-based}}\\
\hspace{1em}Instant-NGP~\cite{muller2022instant}\xspace 
& 22.04 & 0.866 & 0.414
& 19.95 & 0.781 & 0.405
& 18.33 & 0.745 & 0.412\\

\hspace{1em}UC-NeRF~\cite{cheng2023uc}\xspace 
& 26.75 & 0.904 & 0.422
& 26.09 & 0.867 & 0.413
& 27.66 & 0.887 & 0.393\\
\hspace{1em}MARS~\cite{wu2023mars}\xspace 
& 25.53 & 0.871 & 0.185
& 26.97 & 0.868 & 0.158
& 28.04 & 0.870 & 0.155\\
\hspace{1em}NeRFacto~\cite{tancik2023nerfstudio}\xspace 
& 22.41 & 0.862 & 0.367
& 21.51 & 0.817 & 0.326
& 18.33 & 0.745 & 0.412\\
\hspace{1em}EmerNeRF~\cite{yang2023emernerf}\xspace 
& 25.97 & 0.893 & 0.167
& 26.87 & 0.863 & 0.162
& 28.43 & 0.889 & 0.139 \\
\hline
\multicolumn{4}{l}{\small\textit{- Gaussian-based}} \\
\hspace{1em}3D-GS~\cite{kerbl3Dgaussians}\xspace 
& 20.73 & 0.872 & 0.402
& 18.86 & 0.786 & 0.436
& 19.24 & 0.797 & 0.393 \\
\hspace{1em}PVG~\cite{chen2023periodic}\xspace
& 22.40 & 0.898 & 0.338
& 22.56 & 0.860 & 0.331
& 23.96 & 0.873 & 0.317 \\
\hspace{1em}GaussianPro~\cite{cheng2024gaussianpro}\xspace 
& 22.04 & 0.866 & 0.414
& 19.95 & 0.781 & 0.405
& 18.33 & 0.745 & 0.412\\
\hspace{1em}DC-Gaussian~\cite{wang2024dc}\xspace 
& 25.00 & 0.908 & 0.329
& 22.07 & 0.851 & 0.347
& 22.61 & 0.867 & 0.333\\

\hline

\toprule
&\multicolumn{3}{c}{Scene004} & \multicolumn{3}{c}{Scene005} & \multicolumn{3}{c}{Scene006} \\
& $\uparrow$PSNR & $\uparrow$SSIM & $\downarrow$LPIPS & $\uparrow$PSNR & $\uparrow$SSIM & $\downarrow$LPIPS & $\uparrow$PSNR & $\uparrow$SSIM & $\downarrow$LPIPS \\ \hline

\multicolumn{4}{l}{\small\textit{- NeRF-based}}\\
\hspace{1em}Instant-NGP~\cite{muller2022instant}\xspace 
& 23.23 & 0.835 & 0.380
& 19.70 & 0.771 & 0.434
& 23.03 & 0.700 & 0.522\\
\hspace{1em}UC-NeRF~\cite{cheng2023uc}\xspace 
& 27.83 & 0.886 & 0.416
& 26.79 & 0.852 & 0.399
& 26.11 & 0.759 & 0.482\\
\hspace{1em}MARS~\cite{wu2023mars}\xspace 
& 28.38 & 0.888 & 0.177
& 23.90 & 0.818 & 0.227
& 26.10 & 0.768 & 0.256\\
\hspace{1em}NeRFacto~\cite{tancik2023nerfstudio}\xspace 
& 22.38 & 0.815 & 0.382
& 19.04 & 0.756 & 0.427
& 22.53 & 0.681 & 0.446\\
\hspace{1em}EmerNeRF~\cite{yang2023emernerf}\xspace 
& 28.34 & 0.886 & 0.179
& 24.19 & 0.819 & 0.215
& 25.91 & 0.758 & 0.264 \\
\hline

\multicolumn{4}{l}{\small\textit{- Gaussian-based}} \\
\hspace{1em}3D-GS~\cite{kerbl3Dgaussians}\xspace  
& 20.31 & 0.845 & 0.460
& 17.90 & 0.785 & 0.453
& 19.87 & 0.693 & 0.499 \\
\hspace{1em}PVG~\cite{chen2023periodic}\xspace
& 25.11 & 0.897 & 0.365
& 23.14 & 0.840 & 0.352
& 22.95 & 0.724 & 0.458 \\
\hspace{1em}GaussianPro~\cite{cheng2024gaussianpro}\xspace 
& 23.23 & 0.835 & 0.380
& 19.70 & 0.771 & 0.434
& 23.03 & 0.700 & 0.522\\
\hspace{1em}DC-Gaussian~\cite{wang2024dc}\xspace 
& 23.35 & 0.848 & 0.348
& 25.12 & 0.896 & 0.355
& 23.02 & 0.788 & 0.431\\
\bottomrule
\end{tabular}
\end{table*}

\setlength{\tabcolsep}{1pt}
\begin{table*}[!t]
\footnotesize
\centering
\caption{Results on our proposed dataset with the different offsets using \emph{left-front, front, right-front} cameras.}
\label{tab:main_3_cam}
\begin{tabular}{llcccccccccccc}
\toprule 
Method &\multicolumn{3}{c}{\textit{w/o} Offset} & \multicolumn{3}{c}{Offset-1m} & \multicolumn{3}{c}{Offset-2m} & \multicolumn{3}{c}{Offset-4m} \\
& $\uparrow$PSNR & $\uparrow$SSIM & $\downarrow$LPIPS & $\uparrow$PSNR & $\uparrow$SSIM & $\downarrow$LPIPS & $\uparrow$PSNR & $\uparrow$SSIM & $\downarrow$LPIPS & $\uparrow$PSNR & $\uparrow$SSIM & $\downarrow$LPIPS \\ \hline

\multicolumn{4}{l}{\small\textit{- NeRF-based}}\\
\hspace{1em}Instant-NGP~\cite{muller2022instant}\xspace 
& 29.24 & 0.888 & 0.262
& 23.52 & 0.847 & 0.344
& 22.39 & 0.815 & 0.382
& 21.05 & 0.783 & 0.428 \\
\hspace{1em}UC-NeRF~\cite{cheng2023uc}\xspace 
& 33.12 & 0.912 & 0.360
& 30.07 & 0.896 & 0.355
& 28.81 & 0.881 & 0.373
& 26.87 & 0.870 & 0.421 \\
\hspace{1em}MARS~\cite{wu2023mars}\xspace 
& 31.37 & 0.887 & 0.151
& 29.28 & 0.874 & 0.157
& 28.54 & 0.869 & 0.165
& 26.49 & 0.847 & 0.193 \\
\hspace{1em}NeRFacto~\cite{tancik2023nerfstudio}\xspace 
& 29.39 & 0.890 & 0.246
& 22.91 & 0.850 & 0.307 
& 22.26 & 0.809 & 0.348
& 21.03 & 0.779 & 0.393 \\
\hspace{1em}EmerNeRF~\cite{yang2023emernerf}\xspace 
& 31.51 & 0.894 & 0.146
& 29.41 & 0.878 & 0.152
& 28.66 & 0.873 & 0.160
& 26.62 & 0.851 & 0.188 \\
\hline

\multicolumn{4}{l}{\small\textit{- Gaussian-based}} \\
\hspace{1em}3DGS~\cite{kerbl3Dgaussians}\xspace  
& 29.74 & 0.914 & 0.312
& 22.08 & 0.842 & 0.359
& 21.34 & 0.824 & 0.402
& 19.47 & 0.796 & 0.441 \\
\hspace{1em}PVG~\cite{chen2023periodic}\xspace 
& 33.33 & 0.933 & 0.256
& 26.62 & 0.878 & 0.318
& 25.50 & 0.870 & 0.332
& 23.35 & 0.849 & 0.360 \\
\hspace{1em}GaussianPro~\cite{cheng2024gaussianpro}\xspace 
& 28.13 & 0.889 & 0.321
& 21.90 & 0.839 & 0.379
& 20.85 & 0.822 & 0.404
& 19.36 & 0.795 & 0.443 \\
\hspace{1em}DC-Gaussian~\cite{wang2024dc}\xspace
& 30.23 & 0.912 & 0.271
& 26.74 & 0.883 & 0.315
& 25.53 & 0.8715 & 0.329
& 23.53 & 0.860 & 0.357 \\
\bottomrule
\end{tabular}
\end{table*}

\end{document}